# On the universal definition of intelligence

A definition for Human - AI Comparison

Joseph Chen
University of Tokyo

## Abstract

This paper aims to propose a universal definition of intelligence that enables fair and consistent comparison of human and artificial intelligence (AI). With the rapid development of AI technology in recent years, how to compare and evaluate human and AI intelligence has become an important theoretical issue. However, existing definitions of intelligence are anthropocentric and unsuitable for empirical comparison, resulting in a lack of consensus in the research field.

This paper first introduces four criteria for evaluating intelligence definitions based on R. Carnap's methodology of conceptual clarification: similarity to explicandum, exactness, fruitfulness, and simplicity. We then examine six representative definitions: IQ testing, complex problem-solving ability, reward optimization, environmental adaptation, learning efficiency, and predictive ability, and clarify their theoretical strengths and limitations.

The results show that while definitions based on predictive ability have high explanatory power and empirical feasibility, they suffer from an inability to adequately explain the relationship between predictions and behavior/benefits. This paper proposes the Extended Predictive Hypothesis (EPH), which views intelligence as a combination of the ability to accurately predict the future and the ability to benefit from those predictions. Furthermore, by distinguishing predictive ability into spontaneous and reactive predictions and adding the concept of gainability, we present a unified framework for explaining various aspects of intelligence, such as creativity, learning, and future planning. In conclusion, this paper argues that the EPH is the most satisfactory and universal definition for comparing human and AI intelligence.

# Introduction

I believe that the long-lasting debate about how intelligent AI agents are can finally come to an end. For over 600 years, intelligence has denoted a specific concept of mental abilities that underlies intellectual behaviors, including understanding, decision-making, goal achievement, spontaneous improvement, and many more. Although there are no compelling reasons to assume that intelligence exists only in humans, research on intelligence in non-human subjects has been far more difficult than research with human subjects due to communication issues. Therefore, for a long time, research about universal intelligence – intelligence that is not limited to human subjects – has not drawn enough attention. However, this situation was completely reversed by one of the most critical emerging technologies that humans have been increasingly interested in.

Artificial intelligence (AI) is one of the fastest-growing technologies, not only in terms of industry and market growth but also in the size of research and development efforts dedicated to creating better AI – faster ones, smaller ones, and most importantly, more intelligent ones. But what kind of AI can be considered more intelligent? In terms of perception capabilities, AI has already taken a significant step closer to human-like abilities.

In addition to pursuing higher accuracy in existing tasks, scientists have also used multimodal methods in recent years to enable AI to not only receive information from a single external stimulus (as external inputs), but also to process signals from vision, hearing, touch, and other senses simultaneously, just like how most humans do. Some scholars have also sought to expand AI's output modes, enabling it to control mechanical actuators in addition to text output, thereby achieving motor capabilities similar to those of humans. In this article, we refer to AI systems that use multiple modules to achieve motor and cognitive abilities comparable to those of humans as AI agents.

In the past, Dreyfus argued that AI systems cannot possess genuine intelligence, since intelligence requires a body (Dreyfus, 1979). By discussing AI agents, we can assume that AI has access to perception and motor abilities, just like humans, and that the input and output patterns available to it do not limit the intelligence of AI systems. Note that discussing AI agents rather than individual AI algorithms addresses only the difference between AI agents and humans in perception and motor abilities, not intelligence itself. Thus, we still need to discuss whether AI truly possesses intelligence and, if so, to what extent. In the following sections, we will focus on and refer to AI agents as AI unless otherwise specified.

## The need for a universal definition

We currently have an urgent need for a definition of intelligence applicable to AI agents, and we are dissatisfied with the current state of affairs. Why do we believe it is urgent? Because there is a high demand across many fields to know the comparison of intelligence levels between AI agents - AI agents and human-AI agents. If we can accurately make these two comparisons, we can achieve significant breakthroughs in areas such as: (1) decision making; (2) scientific research; and (3) public understanding of AI agents. We can use this definition to answer questions such as:

"Is intelligence artificially achievable?"
"How is artificial intelligence different from non-artificial intelligence?"
"Can we evaluate artificial intelligence like we evaluate human intelligence?"
"How do we know when an AI agent can perform better than a human in a complex task?"

These questions have been raised repeatedly in recent years and stem from the lack of a concrete definition of intelligence, particularly when the scope of intelligence is extended beyond humans to include other seemingly intelligent beings.

Then why do we claim dissatisfaction with the current interpretation of intelligence? While several definitions of intelligence have been proposed over the years to advance scientific research, none are widely agreed upon, and most are not practical for general implementation and testing. This leads to everyone using their own methods of measurement, resulting in the following tragedies frequently occurring:

1. Companies release new AI models that users perceive as less intelligent compared to prior versions. (Sarah, 2025)
2. The media repeatedly uses human metrics to assess AI agent capabilities (e.g., IQ tests/college entrance exams), ignoring the differences between humans and AI agents and thereby misleading the public.
3. Many people believe in the singularity hypothesis, where the relationship between AI and humans changes drastically when the intelligence of AI surpasses humans. Still, no one can clearly explain the width of the gap or the current state of AI relative to humans.

This unsuitability confirms that our understanding of intelligence is highly anthropocentric and that directly applying our existing knowledge of human intelligence to AI agents is infeasible. Therefore, a universal definition of intelligence is necessary, which is also the goal of this work.

The attempt to generalize the definition of intelligence is not something new. Several attempts across different research fields are currently under development. Hutter stated, from a computer science and information-theoretic perspective, that intelligence can be formulated as maximizing some kind of a utility function (Hutter, 2005, p. 3). From a biological perspective, Sternberg pointed out that all intelligent behaviors and abilities involve some form of contribution to biological adaptation (Sternberg, 2019, p. 2). In the field of brain and cognitive science, Knill and Pouget suggested that brains are, in general, Bayesian perceptual systems, which in turn inspired Friston to develop a theory of self-organizing systems based on Bayesian inference, known as the Free-Energy Principle (Knill and Pouget, 2004; Friston, 2010).

In this work, I aim to conduct a comprehensive review of strong candidates for definitions of intelligence and to highlight important aspects of its nature, and potentially to identify a definition that enables the comparisons needed to advance science.

## Chapter 1. Finding a satisfying definition

To address the necessity mentioned in the previous chapter, the goal of this work is to find the most appropriate and satisfactory definition to assess AI agent-AI agent and Human-AI-agent intelligence comparisons reasonably.

Ideally, we also hope that this definition can be used to assess human-to-human intelligence. This is because the concept of intelligence is traditionally closely linked to, and widely recognized in, human-to-human comparisons. Therefore, regardless of how we define intelligence, its use will naturally be analogous to intelligence in human-to-human contexts. In other words, we would expect "intelligence" to have the same meaning in both the context of "this AI agent is more intelligent than that one" and "she is more intelligent than he." If this is not the case, it would result in two distinct concepts being simultaneously referred to as "intelligence."

Using the same term for different concepts is not strictly prohibited, though. For instance, in computer science, when describing the complexity of an algorithm, we often define complexity as the degree to which it uses time or space efficiently. In contrast, when describing the complexity of a Deep Neural Network model, we generally use the number of NN layers as the complexity of an architecture. The same word being interpreted differently in two contexts is incomparable, but both contribute significantly to the scientific progress of their respective fields. Nevertheless, such phenomena will lead to numerous unnecessary misunderstandings and undermine our research motivation.

In previous work, Wang proposed four criteria for evaluating a working definition of intelligence (Wang, 2019). This methodology is adapted from Carnap's *Logical Foundations of Probability* (Carnap, 1951). Carnap faced the need to define "probability" precisely, which serves as a foundation for inductive logic and for empirical confirmation and verification. At the time Carnap published his publication, the proposed definitions for probability were just as chaotic, if not more so. Nevertheless, Carnap's methodology provided a sound logical interpretation of the concept, and will be used as the core of the discussion in this work.

## Chapter 2. The Four Requirements

The purpose of this chapter is to ensure that subsequent discussions on definition comparisons go beyond just mere "preference" and instead are based on a reasonable and rational framework for evaluating and comparing the various candidate definitions. This is crucial for a specific kind of research problem, including defining intelligence.

Defining "intelligence" is very different from defining terms like "gold" or "air." It is not only invisible and intangible, but it is also still debated on how intelligence reflects in the physical world. Carnap analyzed these kinds of questions, such as "how to define intelligence", "how to define causality," and "how to define probability," as having similar properties. He pointed out that the datum (evidence in the real world) and the problem's statement are inexact, and called these problems "explication problems," emphasizing that "Strictly speaking, the question whether the solution is right or wrong makes no good sense because there is no clear-cut answer. The question should rather be whether the proposed solution is satisfactory, whether it is more satisfactory than another one, and the like."(Carnap, 1951, p. 4). Therefore, we hope that this satisfaction will not only be reflected in personal preferences, but also provide a reasonable and rational comparison in the context of scientific progress.

The core idea of Carnap's "explication" is to replace everyday, pre-scientific concepts (explicandum) with concepts that withstand scientific scrutiny in scientific discourse (explicantum). The success of this replacement is not based on the truth or falsehood of the explicantum, but on the extent to which it meets the needs of scientific progress. This paper also adopts this viewpoint, focusing on replacing past concepts with "comparable definitions" rather than on "the only correct answer to what intelligence is." The evaluation criteria for these definitions will use the same four elements proposed by Carnap: (1) similarity to the explicanum, (2) exactness, (3) fruitfulness, and (4) simplicity (Carnap, 1951, page 5).

**Similarity to the explicandum**

Carnap argues that explication is the description of the transformation of an abstract concept into a more logically clear definition suitable for scientific contexts. The pre-scientific defined usage of the word and the concept attached with it is called an "explicandum" (plural form "explicanda"), and the transformed definition that is fruitful for scientific research is called an "explicatum" (plural form "explicata").

The requirement of "similarity to the explicandum" measures how well the definition (explicatum) addresses the pre-scientific defined usage (explicandum) of a scientific term. For example, if we define intelligence as the time required to run 100 meters, even if it performs well on the other three requirements—Exactness, Fruitfulness, and Simplicity—we would still not consider it a good definition because of the low similarity to our original understanding of intelligence, since we usually do not consider being intelligent to mean running fast.

Carnap describes this condition as follows: "The explicatum is to be similar to the explicandum in such a way that, in most cases in which the explicandum has so far been used, the explicatum can be used; however, close similarity is not required, and considerable differences are permitted. (Carnap, 1951, page 7)"

He used an example to illustrate the meaning of this requirement. The prescientific term of "fish" roughly equates to marine life, meaning that whales, dolphins, and seals/sea lions could all be classified under the explicandum fish. There is another concept of fish that does not include aquatic mammals such as whales, dolphins, seals, and sea lions. To avoid confusion, we will call it "Piscis". The emergence of factual discoveries led to the replacement of the explicandum Fish with the explicatum Piscis in scientific discussions. Due to their differences, Piscis cannot completely replace Fish in all situations where Fish was used. However, the use of Piscis means that we no longer rely on the ambiguous concept of Fish in scientific discussions. This shift, which was based on new scientific discoveries from zoology, tells us that a good explicatum might sacrifice some similarity to the explicandum for scientific fruitfulness, but this is only possible if the explicatum can be used in most cases that the explicandum has been used so far. (Carnap 1951, page 6)

When proposing the requirements for explication, Carnap attempted to address the problem of what probability is, a problem that had sparked debate among scholars at the time. He observed that most scholars criticized the shortcomings of one another's explicata, arguing that their own explicatum better defined probability. Carnap had a different idea. Instead of focusing on the explicatum as other researchers did, he believed it was more helpful to focus on the explicandum of each proposal. He discovered that the concept that the user is referring to when using the word "probability" typically falls into several categories, such as "Degree of belief", "Credibility", "Relative frequency", "Degree of partial truth", and many others (Carnap, 1951, p. 24).

These categories can be further organized into only two types of explicanda. The two types were "degree of confirmation" and the "relative frequency in the long run", which he named probability$_1$ and probability$_2$, respectively. He argued that the probability concept in people's minds is actually a misconception composed of these two concepts. Not only are the two not interchangeable, but Carnap also points out that no broader concept can encompass both explicanda simultaneously. Therefore, anyone discussing probability will find their argument facing obstacles and difficulties if they don't clearly specify whether they are describing probability$_1$ or probability$_2$. By clearly analyzing the target explicandum of a definition and evaluating the definition against the explicandum, Carnap paved the way for resolving disputes surrounding probability. From this history, we can see that the similarity of the explicandum is one of the most crucial factors in determining the fundamental validity of a definition.

## Exactness

Carnap's definition of exactness requires that a definition's wording be unambiguous and not have multiple meanings. Instead, we aim for precise terminology in definitions to reduce the risk of ambiguity or circularity. For example, the Cambridge Dictionary defines intelligence as "the ability to understand and learn well, and to form judgments and opinions based on reason" (intelligence). Neither "understand" nor "learn well" in this definition is highly precise. On one hand, "understanding," like intelligence, is a word that scientists haven't definitively defined. On the other hand, "learn well" can have different meanings for different individuals, particularly when comparing humans and AI agents. Carnap holds that good extension requires precise wording to eliminate doubts that the same definition yields different results in different contexts.

Carnap describes exactness as follows: "The characterization of the explicatum, that is, the rules of its use (for instance, in the form of a definition), is to be given in an exact form, so as to introduce the explicatum into a well-connected system of scientific concepts. (Carnap, 1951, page 7)."

Carnap points out that the content of the explicatum is less vague than the explicandum. Therefore, there can not be any complete coincidence between the two. However, less vagueness also means that it can be more accurately connected with knowledge in other fields. Thus, we can also understand the requirement of exactness to ensure that such connections are not broken by the vagueness of the wording. Carnap used the explication of Temperature as an example, explaining temperature in terms of the explicatum of molecular motion, which makes it easier to draw accurate connections with other scientific research fields in physics.

Modern scholars often use the term "precision" to denote the exactness of the vocabulary employed in a definition. Scholars argue that precise wording can also be understood as a narrowing of the concept, meaning that it aims to exclude neighboring concepts that are similar and retain only the words that truly represent the essence of the term being explained (Assur, 2006). If terms such as broad, deep, and complex play a significant role in a definition, that definition may be considered imprecise, as these terms can be interpreted arbitrarily.

In this work, we will also examine whether the vocabulary used in the definition remains precise for two distinct groups: humans and AI agents. Many words have different meanings when used to describe humans and AI agents; therefore, definitions that use such words will refer to different concepts when applied to different kinds of intelligent agents. In such cases, we also say that exactness is compromised. An example is "Memory." If we define intelligence as "the ability to optimize the utility of one's memory," while this can indeed be well measured in both humans and AI agents, the difference in the meaning of the word "memory" between the two may also introduce interpretive ambiguity. We can say "the memory is full" or "add more memory" for AI agents, but we cannot relate the sentences to human cases. In contrast, words that describe similar functional behaviors in humans and AI agents can be considered to retain exactness. For example, "arithmetic calculations."

## Fruitfulness

Carnap describes fruitfulness as follows: "The explicatum is to be a fruitful concept, that is, useful for the formulation of many universal statements (empirical laws in the case of a nonlogical concept, logical theorems in the case of a logical concept). (Carnap, 1951, page 7)." The metric of fruitfulness measures the extent to which an explicantum can inspire new inferences, laws, predictions, and measurements. An explication aims not merely to unify the interpretation of a term, but to make the concept a running cog in the machine of scientific progress. This is the core of the fruitfulness requirement. This paper will maintain this core and use the following operations to determine how satisfactory the fruitfulness of a definition is: (i)testability, and (ii) reflection. Importantly, these sub-items are not the definition of fruitfulness, but rather operations that we use to identify if a definition may be fruitful.

Whether a definition is truly fruitful can only be judged in the future by how many scientific theories connect with it. We can observe and determine which kinds of definitions tend to be more fruitful, but we cannot guarantee that they will actually make substantial contributions to science in the future. This can be seen in Carnap's example of psychological quantifications that are difficult to reproduce. But to show this, we first need to introduce a method for classifying concepts.

Carnap believes that there are three special kinds of concepts in scientific research: classificatory, comparative, and quantitative concepts (Carnap, 1951, page 9). Classificatory concepts categorize things or things into multiple non-overlapping classes, and most pre-scientific concepts fall into this category. Examples include metal or non-metal, fish or non-fish, intelligent or non-intelligent. Comparative concepts (sometimes called topological or order concepts) describe the result of a comparison. Examples include "warmer," "more flexible," or "more intelligent." Quantitative concepts describe numerical attributes derived from measurement or calculation. Carnap provided examples of quantitative concepts such as volume, temperature, IQ, and length of time. Generally, quantitative concepts are more fruitful than comparative concepts, which in turn are more fruitful than classificatory concepts. For instance, the quantitative concept of temperature is highly fruitful, as it occurs in many important laws. However, not all quantitative concepts possess high fruitfulness. Carnap points out that in some psychological studies, "a quantitative concept was defined by an exact description of tests, but that the expectation of finding laws connecting the values thus measured with values of other concepts was not fulfilled; then the concept was finally discarded as not fruitful."(Carnap, 1951, page 14) This is an example of how deviating from the core competence of fruitfulness leads to explicatum being discarded. I will proceed to explain how we will examine the fruitfulness of a definition by (i) testability and (ii) reflection.

First, we will assess the fruitfulness of intelligence with respect to its testability. For example, using the total amount of electrical neuroactivity in the brain over a specific period as a measure of intelligence is difficult to test with current technology. A definition of intelligence that is highly testable must not only include a reasonable testing method but also consider whether the method is cost-effective and acceptable to researchers. If possible testing methods of a definition require excessive labor or cost that researchers of intelligence might not be able to handle, we might conclude that this definition lacks fruitfulness.

Secondly, we can judge this by **how well the test results reflect real-world outcomes**. When we say we want to measure and compare human-AI agent intelligence fairly, it means we expect to observe a close correlation between the measured results and the actual performance in the real world. Because testing environments cannot encompass all possible real-world conditions, bias arising from environmental differences is inevitable. For some definitions, the results measured in the testing environment may differ significantly from actual performance. Therefore, the risk of accepting such a definition increases, and its scientific validity becomes questionable. Formally describing this principle, we can describe the reflectness principle as: If, according to theory T, Subject S performs better than S* in the testing environment, then T reflects well if (other things being equal) S is also likely to perform better than S* in similar real-world environments.

## Simplicity

Carnap describes fruitfulness as follows: "The explicatum should be as simple as possible; this means as simple as the more important requirements (1), (2), and (3) permit. (Carnap, 1951, page 7)" Later philosophers also paid some attention to this concept. Philosophers subdivided the concept of simplicity into many types. For example, elegance, which means the syntactic simplicity of the principle, and parsimony, which means the ontological simplicity, or kinds of entities that a theory commits to (Baker, 2022). Carnap also stated that "The simplicity of a concept may be measured, in the first place, by the simplicity of the form of its definition and, second, by the simplicity of the forms of the laws connecting it with other concepts. (Carnap 1951, page 7)"

In the section on exactness, we argued that a good definition must use precise terms to clarify its scope. However, this can encourage excessive verbal description. Elegance aims to prevent such over-evaluation. Regarding the evaluation method of parsimony, we can refer to Quine's *ontological commitment*:

> *If two theories, $T_1$ and $T_2$, have the same ontological commitments except that $T_2$ is ontologically committed to Fs and $T_1$ is not, then $T_1$ is more parsimonious than $T_2$.* (Bricker, 2014)

However, this requirement is believed to have lower priority than the others. Carnap stated that "This property, however, is only of secondary importance. Many complicated concepts are introduced by scientists and turn out to be very useful. In general, simplicity comes into consideration only in a case where there is a question of choice among several concepts which achieve about the same and seem to be equally fruitful; if these concepts show a marked difference in the degree of simplicity, the scientist will, as a rule, prefer the simplest of them."(Carnap, 1951, page 7).

## Requirement trade-offs

It's worth noting that there's a trade-off between the four requirements mentioned above. Exactness and simplicity, as previously mentioned, are just prominent examples; in fact, all requirements are mutually restrictive to some extent. For instance, if we define intelligence as the speed of solving 1,000 math equations, although we have a simple and exact definition, we are sacrificing the definition's simplicity to the explicandum, as it is far from our folk concept of intelligence.

Finally, using the tools described above, we conclude that a definition should not be significantly deficient **in any of the four** requirements proposed by Carnap. Such a definition would be considered inadequate. Instead, we will address the trade-off among the requirements while exploring which definition best balances them, thereby yielding the most satisfactory definition. In the next section, we will attempt to verify these requirements against prior work that may provide a useful definition of intelligence for both humans and AI.

# Chapter 3. Prior studies

## Section 3.1 Definition #1: Intelligence is what IQ tests test

### Section 3.1.1 Introduction

Well known as the opinion that Edwin G. Boring defends in his work, "Intelligence as the tests test it", the concept that intelligence is what is being tested in IQ tests is also known as an operational definition of intelligence (Boring, 1961). An operational definition is a kind of definition that is based on an operation, or "the performance which we execute to make known a concept" (Stevens, 1935, pp. 323–324, 330).

Van der Maas et al., in their publication "Intelligence Is What the Intelligence Test Measures. Seriously", stated that "the core of intelligence research rests on the positive manifold: the fact that all intelligence subtests, ranging from scholastic tests to tests of social intelligence, correlate positively." (Van der Maas et al. 2014, page 12). It is the strong correlation between intelligence tests and academic achievement that has given IQ test scores practical utility and strong explanatory power, making them an irreplaceable part of psychology, law, and education. Substantial research worldwide provides evidence of correlations among battery subtests and between test scores and real-world performance, suggesting the existence of measurable intelligence.

When a test taker takes a subtest, the result of each test item indicates only their level of ability on that item. This level of ability consists of two components. One is what Boring called the common factor, and the other is those dedicated abilities that only appear in that specific subtest and do not affect the scores of other subtests, referred to as "special ability".

Boring stated, "Thus we see that there is no such thing as a test for pure intelligence. Intelligence is not demonstrable except in connection with some special ability. It would never have been thought of as a separate entity had it not seemed that very different mental abilities had something in common, a 'common factor.'"(Boring, 1961, p. 2). Boring believed that by summing up the scores of the test items, the "special ability" that is not intelligence will cancel out each other, revealing what Boring called accurate intelligence, which he believed is the common factor.

We can understand this view of intelligence as emphasizing a definition's scientific fruitfulness. Defining intelligence as what the IQ tests test is, quoting Boring's words, "the only point of departure for a rigorous discussion of the tests" (Boring, 1961, page 1). We anticipate that this definition, which demonstrates a high degree of scientific fruitfulness in human intelligence, might also show considerable fruitfulness in measuring AI agents' intelligence.

Section 3.1.2 IQ test scores as a universal definition of intelligence

**Similarity to the explicandum**
First, we must examine the implications of treating IQ scores as a definition of intelligence when applied to AI agents. While many other definitions derived from human observation are challenging to interpret into how AI can be measured, this one is relatively straightforward. We can treat AI agents as IQ-test takers, and let their scores determine their intelligence. The higher the score, the higher the AI agent's intelligence. Furthermore, test subjects that achieve similar results, whether human or AI agents, are considered to possess similar levels of intelligence. In other words, an AI agent that scored 130 on an IQ test will have a comparable level of intelligence to a human who also scored 130.

It's worth noting that IQ test scores are calculated using a normalization method based on the distribution of human test-takers' scores. This means that, from an individual's IQ test score, we can determine how many standard deviations they deviate from the mean. This normalization process is effective when only humans are tested. However, when testing AI agents, we must consider the possibility that the distribution of intelligence between AI agents and humans differs. If the AI agent's intelligence distribution differs from that of humans, a situation may arise in which the AI's context differs significantly from that of humans. Yet, the distribution appears concentrated in a human-centric context. For example, raising an AI agent's IQ test score from 50 to 55 might be a significant improvement, even if it doesn't appear much different on a human's scale.

Next, we will discuss some challenges regarding the requirement of similarity to the explicandum. When using IQ scores to explain human intelligence, this definition has been criticized by many scholars for failing to genuinely reflect the explicandum of intelligence we perceive in our daily lives (Ganuthula & Sinha, 2019, p. 2). Their criticisms include: (1) IQ tests ignore many cognitive abilities that are difficult to measure, for example, creativity or emotional sensitivity; (2) IQ tests prioritize crystallized intelligence over fluid intelligence, which means that while prior knowledge is valued, learning abilities get less attention; (3) Some people might perform well on one IQ test but poorly on another, indicating that the test maker and the environment greatly influence test results. Although these are mostly criticisms of using this definition to measure human intelligence, these problems will inevitably be preserved or even amplified when the definition of intelligence is extended to include AI agents.

Beyond the measurement of human intelligence, the scope of interpretability in this definition also poses a major issue. Suppose we define an individual's intelligence as the score from an IQ test in pursuit of fruitful numerical indicators for measurement and comparison. By doing so, we also drastically narrow the scope of what constitutes intelligence. Not only are animals such as cats and dogs excluded from being considered intelligent, but so are infants and individuals who are unable to take the test due to illiteracy or communication difficulties. Boring had pointed out that "this is a narrow definition" when proposing to use IQ test scores as a definition of intelligence (Boring 1961, page 1). Still, he was willing to trade some similarity to explicandum for the sake of scientific fruitfulness.

A weaker argument attempts to address this problem, stating that IQ test results are merely a valid approximation of pure intelligence; thus, those unable to take the test simply lack this approximation (defined as intelligence), but do not necessarily lack pure intelligence. These kinds of claims aim to trade back partial fruitfulness to solve the undersatisfying similarity to the explicandum. However, even if we consider this a weaker argument, problems with the scope of the definition remain.

For instance, we cannot use "the ability to take an IQ test" as a threshold for intelligence; that is, we cannot assume that any individual capable of taking an IQ test possesses a higher level of intelligence than those who are unable to take it. This is because taking an IQ test requires communication with others. We can imagine an individual who, due to limitations such as communication difficulties, is unable to take the test yet possesses a higher level of intelligence than someone who can take the test but obtains a low ability score. Therefore, IQ test results can only serve as a comparative standard among subjects who can take the test normally and cannot reveal the relative level of intelligence among subjects who are unable to take the test.

Another good example showing this lack of similarity is the Flynn effect. Flynn found that, overall, IQ test scores increased year by year (Flynn, 1987). This growth rate is substantial, approaching the average rate of 5-25 points per generation. Many scholars believe that this growth rate exceeds the actual increase in human intelligence. In other words, they do not believe that IQ test scores accurately reflect intelligence, because the intelligence we perceive does not show such significant differences between generations (Rushton and Jensen, 2010).

Furthermore, because IQ test scores are computed from the mean and standard deviation, test developers must frequently adjust item difficulty or recalculate the mean and standard deviation. This has led some to question his fairness, arguing that results measured ten years ago should not differ from those measured today. Proponents of using IQ tests as a definition of intelligence have yet to provide an explanation that adequately accounts for the Flynn effect, leaving lingering doubts in this area.

**Exactness**

Exactness for this definition, as in wording precision, is generally considered good. However, some researchers argue that IQ tests are not well-suited to AI agents and advocate for developing IQ tests specifically for AI assessment, which might, in turn, affect the accuracy of the definition (Hutson, 2025; Shi, 2022). Creating IQ tests specifically for AI can not only refine observational items and re-examine many questions considered too easy for AI (such as knowledge-based questions), but also correct the distribution of intelligence levels. In other words, the level of human intelligence can be inferred from the distribution of human test scores, and the level of intelligence of AI agents can be inferred from the distribution of AI agent test scores, allowing both to better understand their position within their respective groups. This is the advantage of doing so.

The drawback of this approach is that IQ test scores can have two distinct meanings: one is the result of a new AI-specific IQ test, and the other is the result of a human-administered IQ test. Because the same term can have different interpretations when applied to humans and AI agents, this definition may be criticized for imprecision, thereby reducing its usefulness for a fair comparison of the intelligence levels of humans and AI agents.

**Fruitfulness**

The use of IQ scores to assess an individual's intelligence has undoubtedly played a crucial role in the history of psychology. It is widely used in psychological research, education, policymaking, and law. It is also widely recognized by the public. Although many cannot accurately explain the different types of IQ test items and the meaning of scores, most people know that the average IQ score is approximately 100 and that higher scores indicate greater cognitive ability. However, in contrast to the evaluation of human intelligence, which is proven to be very fruitful, the fruitfulness of using IQ test scores to measure the intelligence of an AI agent remains highly debatable. Scientists question whether an AI agent that achieves a high IQ test score can also demonstrate high intelligence in areas outside of the IQ test, in other words, whether the tests reflect the abilities in real environments.

There are reasons for such skepticism. One reason is that AI's specialization capabilities far surpass those of humans. Compared with humans, there is a much greater likelihood that an AI will excel on IQ tests but fail to replicate those results in real-world environments. In fact, scientists have already created AI specifically designed to solve IQ test problems (Mandziuk and Zychowski, 2019). From a tool perspective, this ease of specialization is very convenient for users. However, the specialization capability also makes assessing a subject's overall abilities highly difficult. This is a significant flaw in our previously mentioned fruitfulness metric: how well the results in test environments reflect those in the real world. Researchers pointed out that, for scientists aiming to develop highly intelligent systems, using the IQ test score as the only metric might encourage them to adopt a divide-and-conquer approach, meaning that they train AI systems on individual subtests rather than on the generalization ability, to achieve a higher score (Wang, 2019).

Another reason concerns the applicability of this definition to non-human species, specifically to AI agents. First, the theoretical foundation of IQ tests is primarily grounded in human-centric research. It is difficult to determine whether non-human entities, if intelligent, would exhibit the same correlation with IQ tests as humans. Some proponents might argue that newer AI agents tend to have higher IQ test scores, supporting the claim's applicability that IQ tests can be used to measure the progress of AI science. Indeed, experimental results have confirmed this inference (Jaš, n.d.). However, given current training methods for AI models, we cannot rule out a positive correlation between the *training data* used by these agents and the content of the IQ tests they took.

Suppose that there are two trained AI agents, A and B. If the training data that A used contains contents related to the IQ test they take for measuring their intelligence, while the training data B does not, then there is a chance that A will perform better than B in the IQ tests, even if in reality B is superior in other abilities. The training data used to train AI systems nowadays is so large in volume that it is very hard to determine whether there are excessive correlations between the training data and the content of IQ tests. If this positive correlation exists, even if AI agents deviate further from true intelligence, models trained with more data may still achieve higher IQ test scores. This undermines the contribution of this definition to scientific progress.

## Simplicity

On the surface, using IQ score as a definition of intelligence seems to satisfy the principle of simplicity. However, in reality, it's quite the opposite case. We can observe this in the structure of many current IQ tests. Generally, an IQ test comprises several subtests, each with numerous items. The final score is calculated as a weighted average of the subtests, similar to how scores are calculated in decathlons, and then normalized to the scores of other test takers.

This is a relatively complex operation. For researchers, fully understanding the meaning of IQ tests requires a detailed examination of the capabilities measured in each subtest and their respective weights. AI system developers are encouraged to adopt a divide-and-conquer approach, attempting to achieve IQ test scores that exceed the actual capabilities of AI systems in the real world.

Regarding the parsimony evaluation of this definition, we note that it rests on numerous specific premises. For example, premise (1): Special biases in subtests cancel each other out. These special biases are those other than $g$ that affect the subject's performance on subtasks. In reality, this canceling-out effect varies from test to test, depending on the author, the environment, and the test taker. Premise (2): The scoring system is accurate and reflects real-world scenarios. Premise (3): The test-taking environment doesn't affect the subject. This has led to IQ tests being overly complex.

### Section 3.1.3 Conclusion

The operational definition of intelligence - "intelligence is what the IQ tests test" is highly valued and widely used in many countries, due to it being well-defined, practically testable, and having a strong correlation to attributes like academic achievements. It is widely known in anthropocentric contexts and has proven highly fruitful for psychological research.

From the above, we also discussed how this definition has the fruitfulness advantage of being highly testable and with strong prior influence, but as a universal definition, it is inferior in that it does not simultaneously ensure the test-reality reflection (fruitfulness) and sufficient similarity to the explicandum. These problems are especially apparent when considering non-human subjects, so additional effort will be required to ensure the definition is accepted as universal. It is also worth noting that half a century had already passed since Boring proposed this view of intelligence. As Boring stated, "measurable intelligence is simply what the tests of intelligence test, until further scientific observation allows us to extend the definition", since new research has been conducted, it is possible that we find hints on how to extend this definition elsewhere (Boring, 1961, p. 1). In the following sections, we discuss definitions derived from recent scientific observations.

## Section 3.2 Definition #2: Intelligence as Complex Problem Solving

### Section 3.2.1 Introduction

The idea of intelligence as the ability to solve complex problems has been widely used across fields, including psychology, animal cognition, computer science, and education (Lenat & Feigenbaum, 1991; Minsky, 1987; Resnick & Glaser, 1975; Wenke et al., 2004). Researchers supporting this view emphasize the importance of developing methods that are not visible to the agent during learning to enable problem-solving (Griffin & Guez, 2014). For example, in educational contexts, problem-solving abilities are recognized as a cross-curricular skill with high real-world relevance (Greiff et al., 2013). Sternberg proposed the concept of *successful intelligence*, defined as: "the use of an integrated set of abilities needed to attain success in life, however an individual defines it, within his or her sociocultural context."(Sternberg and Grigorenko, 2003, page 208).

Scientific observations were made by administering IQ and CPS (complex problem-solving) tests to the same samples, and the results showed moderate-to-high positive correlations. A meta-analysis of these experiments concluded that the overall correlation between the two rests at 0.43 (Stadler et al., 2015, p. 1). Another strong motivation supporting this definition is that there are mappings between the capacities for complex problem-solving and both fluid intelligence (Gf) and crystallized intelligence (Gc). In Hambrick's work, it is stated as "There is a general capacity for solving novel problems that map onto Gf/Intelligence A, and there is knowledge acquired through the exercise of this capacity that maps onto Gc/Intelligence B."(Hambrick et al., 2019, page 560). The Intelligence A and Intelligence B stated in the previous quote refer to fluid and crystallized intelligence, respectively. From the above observations, we can see that defining intelligence as complex problem-solving abilities has considerable explanatory power in human intelligence.

Extending the definition of intelligence as complex problem-solving abilities to non-human intelligence is also relatively straightforward. In fact, we have seen this concept in practice as a method of measuring the cognitive abilities of animals (Rosenberg et al., 2021). Scientists have meticulously designed numerous experiments that control for variables (such as prior experience) to study the problem-solving abilities of birds and dogs (Shaw, 2017; Kaplan, 2015; Elliot and Scott, 1965). Furthermore, meta-analyses have highlighted measurement methods that focus on individual animals rather than groups (Rowell et al., 2021).

Section 3.2.2 Complex problem solving as a universal definition of intelligence

**Similarity to the explicandum**

We say that AI scientists are constantly striving to develop AI with greater intelligence, but what metrics are they currently using to judge intelligence? One of the most widely used methods is to check against benchmark datasets. These benchmark datasets can be viewed as complex test problems designed for specific problems, and are often samples from a set of real-world problems that scientists intend to solve. These AI problems presented in benchmark datasets are usually complex problems that require more than trivial reasoning to solve. Therefore, we can see that defining intelligence as the ability to solve complex problems is closely aligned with the current development process in AI research.

Current AI research generally holds that higher benchmark scores correlate with better AI performance or greater commercial value (Chollet, 2019). At a glance, this belief seems valid if the real-world tasks of these AI agents highly overlap with the benchmark scope. However, there are circumstances in which this correlation may disappear when the tasks deviate from the original training objective. A well-known example is benchmark leakage. Since both the training and test data in the benchmark come from the internet, AI systems might perform better on the benchmark but not as well in the real world because they have seen similar data in their training dataset (Xu et al., 2024). Another situation concerns the behavior of LLMs, which typically treat text translation as their primary training objective (the problem to be solved), but have shown less appealing performance on mathematical problems (Williams & Huckle, 2024). This issue has led some scholars to suspect that benchmarks actually constrain AI training. Rather than developing more general capabilities, the focus of AI research narrowed down to learning to predict specific benchmarks. In other words, an AI agent that can effectively solve specific complex problems might receive a high rating on this definition, but its performance in general abilities might actually be less impressive. Since we expect intelligence to be the ability to solve problems in general, measuring only the definition of some specific complex problems means that we are measuring something different from our intelligence explicandum.

This observation echoes a phenomenon that many AI researchers have termed the narrow and general dilemma (Ang et al., 2023; Gamez, 2025, p. 8). The narrow and general dilemma refers to the observation that, although many AI agents can achieve excellent results on specific tasks, they still cannot match humans in the breadth of problems they can solve. For example, DeepBlue and AlphaGo, which defeated the best human chess and Go players, are incapable of solving problems outside of the board.

Many researchers believe that one essential condition for intelligence is the ability to solve a *wide* range of problems (Legg and Hutter 2007, page 402). Although many commercially available AI systems currently surpass human performance on specific benchmark problems, they are more often treated as "specialized agents" rather than "general agents" such as humans or animals. Therefore, to measure intelligence in AI agents based on their complex problem-solving abilities, it is necessary to test agents on a wide range of problems, rather than relying on a single benchmark or a set of similar ones.

The narrow and general dilemma raises the following questions about the similarity to the explicandum of this definition: when we need to decide between an agent that performs very well on one problem and another that performs well on a wide range of problems, which one has more intelligence? How can we use this definition to explain AI systems that perform better at complex problems than on simple problems, which is unlikely to happen with human subjects?

Another criticism of complex problem-solving abilities is that they are highly dependent on motor skills, whereas we argue that intelligence is primarily a mental ability. I will invite you to a thought experiment I call the man-in-robot scenario. Imagine a normal person sitting in a robot with superhuman strength, incredible speed, and extreme motor precision. In any problem that needs to be solved, whether simple or complex, the man in the robot will have an advantage over the same person without the robot. However, our understanding of intelligence suggests that both the person and the "person in the robot" system should possess a similar level of intelligence, since it is the person controlling the robot. This is not the case when intelligence is defined as complex problem-solving, and therefore considered a flaw when it comes to the similarity to the explicandum for this definition.

**Exactness**

The lack of "exactness" is the most significant flaw in this definition. Proponents of this definition argue that "complexity" is a crucial element in determining the problems of interest. They argued that "simple problems" are insufficient to distinguish among intelligence levels in AI agents or humans. The inability to solve simple problems, such as a maze or a puzzle box, is an inadequate discriminative measure of intelligence. For example, mice can navigate a maze in a split second; specific slime molds that can find the shortest path in a maze also exist, so both living beings possess some degree of problem-solving ability (Adamatzky, 2011). Even if these organisms are better at solving these problems than humans (e.g., by finding a solution in less time), we would not conclude that their intelligence surpasses that of humans. Boring argues: "The test of intelligence in an animal is a maze to learn or a puzzle box to open. Such a performance measures a special ability along with the 'common factor,' and it cannot be considered a test of intelligence."(Boring, 1961, page 36).

Another example of simple problems for humans, such as how to use scissors, requires little cognitive effort for most people and thus seems far from a fruitful measure of intelligence. Therefore, we can conclude that the element of "complexity" is crucial for this definition (Dörner & Funke, 2017, p. 2). But how exactly can "complexity" be defined? This is the crux of the problem. In general, task complexity varies among test takers, and no single definition can be used to quantify it precisely. Because there's no universally accepted standard for defining complexity, experiments are often based on open real-life problems encountered in human life, such as "achieving a certain revenue in a shop over a period of time." These kinds of problems are tested in simulated worlds called "microworlds", where test takers are put in a controlled simulated environment and every decision and outcome can be traced by observing researchers.

However, even these real-world problems may pose different difficulties for individuals, particularly when applied to AI agents, where hardware differences can be substantial. Therefore, we can see that limiting the set of problems that we are interested in to "complex problems" will weaken the precision of the definition, as imprecise words like "complex" may have different meanings, not only between humans and AI agents, but also may be ambiguous between individuals of the same species.

**Fruitfulness**

Some researchers criticize this definition, arguing that the more complex the problem, the less likely it is to have a single correct answer, and that it's even questionable whether any answer is accurate (Vamplew et al., 2021, p. 12). For example, consider the issue of ethics. If an AI agent achieves high sales through deception or other unethical means, yet is unaware of the consequences and ethical implications of such behavior, can we reasonably rate its intelligence level as high? This also indicates that the testability of this definition declines as the subjects become more intelligent. This is because, in many cases, measuring and comparing higher intelligence requires subjects to solve more complicated problems. Imagine that AI science progressed so much that the complex problems humans posed were quickly optimized, and that, by the answer alone, we could not tell which of the two AI agents was more intelligent. We will need to rely on even more complex problems for which humans do not know the correct answers. This is why we can claim that the definition's fruitfulness with respect to testability is problematic regarding highly intelligent subjects.

Regarding the ability to reflect from test data to real-world environments, it is well known that AI agents, as a result of machine learning processes, perform less optimally when faced with data from entirely unseen categories. If this definition of intelligence were to be adopted, would the evaluation of AI agents' intelligence be biased by the testing distribution? Imagine an AI agent trained to predict for a bank whether to lend money to candidates. Assume that while it performs extraordinarily well on previously seen data, like 25~40 years old Japanese men, its predictions become nearly random guessing when unseen data, like a 50-year-old American, appears. If the distribution of seen/unseen data decides the AI agent's performance outcome, is it reasonable to evaluate the intelligence of the AI agent by only looking at the results in the testing environment? What type of data distribution would be appropriate for evaluating the system's complex problem-solving abilities? This adds to the fruitfulness issue of this definition.

**Simplicity**

This definition perfectly satisfies the condition of simplicity. Unlike IQ test scores, which depend on complex formulas, problem-solving abilities are straightforward, intuitive, and readily understood by almost everyone. The reason is that this definition of intelligence is strongly motivated by biological considerations. Beings with higher problem-solving abilities tend to have a higher survival rate. This also aligns with our desire to increase intelligence and with the common belief that it is better to have higher intelligence than lower.

Section 3.2.3 Conclusion

Defining intelligence as the ability to solve complex problems has been fruitful in fields such as education and AI science. This definition is closely aligned with the current paradigm of measuring AI research progress through benchmarks. The primary issue is the precision of this definition: the complexity of the proposed issues may vary across subjects, so what counts as complex is not exact, even though complexity is a crucial element in their definition. The design of testable problems also poses a challenge for the definition's universality, as it involves a narrow-general dilemma: whether breadth intelligence, which solves many problems, should be considered more intelligent than deep intelligence, which solves few problems but finds optimal solutions. The testability of this definition depends on the range of problems encountered during a testing session, and it may encounter obstacles when addressing more complex problems that lack exact answers.

Regarding relationships with other definitions, we note that this definition includes the one in which intelligence is defined as what IQ tests measure. This can be done by noting that solving IQ test problems is a subset (or special case) of solving complex problems. It seems we can also make the same claim for our next definition, which defines intelligence as optimizing rewards, arguing that reward optimization is just one kind of complex problem. However, compared to this definition, optimizing rewards not only places more emphasis on optimization than solving for an answer, but also, compared to the complexity and variety of complex problems that are used to define intelligence in this definition, the next definition places greater emphasis on the value of rewards as a single indicator.

# Section 3.3 Definition #3: Intelligence as Optimizing Rewards

### Section 3.3.1 Introduction

In the previous definition, we discussed defining intelligence as the ability to solve complex problems. One issue we encountered in that definition was the need to address the breadth of intelligence with a wide range of problems. Another group of scientists argues that this complex interaction between abilities and desires can be reduced to a single, generic objective: maximizing reward. (Silver et al., 2021, p. 1). For example, let's say a squirrel hid its nuts in a hole. We can interpret this behavior as the coordinated activity of multiple subsystems, including motor control, memory, perception, knowledge, and social intelligence, to minimize the squirrel's hunger (Silver et al., 2021, p. 2).

However, some scholars have argued that a single scalar reward cannot accurately capture intelligence. They have observed that many situations in human intellectual behavior involving reward optimization cannot be explained by a single scalar reward (Vamplew et al., 2021). For example, appetite and the desire to sleep are two desires that coexist and are irreplaceable. At a more fundamental level, in addition to dopamine, which has the most significant impact, mechanisms influencing attention and reward processing in the human brain also include neurotransmitters such as serotonin and endorphins. Therefore, if we want to match the complexity of the human reward system, a single scalar reward is inevitably insufficient for AI agents. However, designing a non-scalar reward is complex, and different reward types pose distinct problems, as we will show later. To simplify the discussion, we will divide rewards into two categories: external and internal rewards. External rewards are those provided from outside the subject, such as by scientists, companies, or the environment itself. They often take physical forms, such as treats used in animal intelligence experiments. By contrast, internal rewards originate within the subject and are typically biological, such as dopamine secretion.

**External rewards**

Using maximizing external reward as a definition of intelligence presents two significant problems. The first problem can be observed through Stanford University's marshmallow experiment. This experiment involved giving children a certain number of delicious marshmallows. However, if the children decided to delay receiving their marshmallows, they could receive more. The phenomenon that researchers aim to observe is called delayed gratification. Humans often consider delayed gratification to be more intelligent than greedy behaviors. In the long run, being willing to postpone receiving the reward (the marshmallow) yields greater rewards and is therefore evaluated as indicating higher intelligence (Mischel, 2014).

The problem is that when reward maximization is used as the definition of intelligence, it leads to the overvaluation of greedy agents, contradicting the general perspective that appreciates delayed gratification. This overvaluation arises because greedy agents are always considered more intelligent, as they maximize their reward at every stage.

This undermines the importance of planning beforehand, which is another ability that should be regarded as an aspect of intelligence.

What if we don't observe the subject's reward at the moment each action is decided, but rather after the event's outcome (i.e., after the delay of delayed gratification)? Then we encounter the second problem: even if the result is better, we have no way of knowing whether the subject foresaw the outcome at the moment of the action, or whether the outcome just happened to be a better choice. In other words, the causal relationship between the action and the conclusion becomes weak. Taking the marshmallow experiment as an example, even if the subject ultimately received more marshmallows by delaying, we don't know whether this was because they foresaw the outcome or because they desired the marshmallows immediately but made the wrong choice. Thus, this method also undermines the importance of making optimal decisions at every given point. This causation problem exists not only in long-term measurements but also in short-term methods that measure the reward based on the subject's state after each action.

However, compared with long-term measurements that assess only the gain in the final state, short-term measurements, because they produce assessments at every stage, provide greater causal relationships and more evidence that the subject can optimize rewards. This is similar to how people are more confident in entrusting their money to investment advisors who can consistently generate returns over short periods, compared with investment advisors whose investments experience significant gains and losses, even if, at some point, the latter outperform the former.

Finally, defining intelligence in terms of extrinsic rewards also raises the issue of personal preference. Individuals will have different preferences for extrinsic rewards. For example, a child who does not like marshmallows may be an outlier in the Marshmallow experiment. If the same reward in the same testing environment motivates different individuals to varying degrees, intelligence cannot be compared fairly. To conclude, defining intelligence as the maximization of extrinsic rewards presents numerous difficulties.

**Internal rewards**

If external rewards don't work, what about internal rewards? Scholars have proposed a theory about how the human brain processes rewards, known as the free energy principle (FEP), that may serve as a reliable internal reward framework. This principle, famously proposed by Friston, states that any self-organizing system at equilibrium with its environment must minimize its free energy (Friston, 2010). FEP is widely regarded as a suitable model for internal reward mechanisms. One of the core concepts of this hypothesis is surprise suppression, meaning that the brain works to minimize free energy (which is the variational upper bound of surprisal) in the system. This biological phenomenon motivates intelligent living beings to continuously pursue a better environment with lower surprise, thereby remaining in it and gaining a greater survival advantage. Here, surprise refers to the variance that external factors introduce into the system, or the negative logarithm of the probability of an event.

The emergence of the FEP explains how the brain infers causation from received perceptual signals. It also describes how beliefs are updated using Bayesian updates over probability distributions. According to FEP, all intelligent behaviors can be understood as the pursuit of minimizing surprise. This means that the better an agent minimizes surprise, the greater the reward it receives. This not only makes a single reward feasible but also underscores the widely held view that the subject must interact with the environment to exhibit intelligent behavior.

Similarly, the concept of internal reward maximization is also popular in computer science, particularly among researchers studying reinforcement learning (RL) and deep reinforcement learning. RL is a type of machine learning method that focuses on maximizing a reward signal or minimizing the upper bound of a loss function in a dynamic environment. RL agents leverage the fact that they can learn through repeated trial-and-error, requiring only a fraction of the time of real-world agents. Even recently, in subfields of robotic studies such as drone balancing or humanoid robot walking, there remains a wide range of uses in maximizing rewards through RL training tasks.

Section 3.3.2 Reward maximization as a universal definition of intelligence

**Similarity to the explicandum**

We have already discussed the conflict between external rewards and the existing human intelligence explicandum. Next, we will introduce a problem that arises when intelligence is defined by maximizing internal rewards or by the FEP. The Dark Room Problem assumes an environment with minimal information (Piekarski, 2021, page 31). In such an environment, all systems will not receive external stimuli, and no surprises will occur. Scholars argue that if FEP is true, there is no reason for organisms, including humans, to leave the dark room, since doing so increases surprisal. One possible hypothesis is that the course of human evolution led humans to leave dark rooms, thereby conferring a significant survival advantage. Currently, we are uncertain whether this hypothesis is correct. Still, if we define intelligence as maximizing internal rewards, we will encounter problems similar to the dark room problem when applying this definition to AI agents.

We suppose that different states (including the environment and the internal state of AI agents) offer varying levels of rewards to the AI agent. Therefore, a knowledgeable AI agent, upon discovering a state that yields high rewards, will attempt to stay in that state to maximize its rewards. This high-reward state is the AI agent's "dark room." For a practical example, suppose an AI agent is specifically trained to do laundry and designed to earn high internal rewards by cleaning dirty clothes. This AI agent might then deliberately soil the clean clothes to keep washing them and maximize its rewards. A similar scenario is the paperclip problem described by Bostrom, in which an AI agent goes wild in order to generate paperclips as much as possible (Bostrom, 2014). Bostrom argues that if AI can reach a level of intelligence that surpasses humans, no matter how we design our reward functions, it will find ways to exploit design weaknesses to maximize reward. Clearly, this behavior of not considering the consequences of factors other than the reward differs from our intelligence explicandum.

**Exactness**

To make this definition more precise, Silver provides clear mathematical notation for the concepts of agent, system, environment, action, observation, and reward (Silver et al. 2021). However, he did not address the difference in the meaning of reward between AI agents and humans. For example, as noted in the introduction, the human brain uses multiple hormones and neural circuits to process rewards. These desires (such as appetite and sleep) are mutually irreplaceable, and this relationship is not captured by current AI architectures. Another difference is that current AI agents generally stop updating rewards after the training phase. In other words, unlike the human brain, which continuously updates the importance of rewards, many current AI agents freeze the importance of rewards during training and repeatedly execute learned behaviors when the rewards are frozen.

**Fruitfulness**

This definition also hides a testability issue in human-AI agent comparison. Even if we successfully apply this definition to AI agents, thereby enabling fruitful comparisons between AI agents, various problems will still arise when using it to compare humans with AI.

We know that for AI agent-AI agent comparison, we can use the same reward to compare the intelligence of two AI agents. This can also be done in human-human comparison, for example, comparing the profits of two stock operators after a trading day. However, how do we compare the rewards for human-AI agent comparison? Herein lies a dilemma. If we use external rewards for comparison, results may vary across individuals depending on how vital the reward is to them. For instance, we aim to use puzzle games to measure differences in intelligence between humans and AI agents. From the AI agent's perspective, they are likely designed to aim solely for the high score. On the human side, however, enjoyment of the game might be the more essential factor than the score, leading human subjects to compromise some score optimality for greater enjoyment. Therefore, maximizing scores (external rewards) is not comparable across the two sides in this intelligence-testing method.

If we choose to use internal rewards instead, measurement difficulties arise. It is possible for AI agents to analyze their reward accumulation at any given time, but it is difficult for humans to quantify the amount of their internal rewards. Furthermore, we not only need a way to test human internal rewards but also a means to compare internal rewards in two forms, which poses a major problem for testability.

**Simplicity**

This definition is among the best definitions in terms of simplicity.

Intelligence, defined as the ability to maximize rewards, is a simple, intuitive, and neuroscience-based concept that has been fruitful in AI science, particularly in reinforcement learning. However, there are critical issues regarding its similarity to the explicandum and its fruitfulness.

The intuitive understanding of this definition, when framed in terms of external rewards, is biased toward selfishness and short-sightedness. It was also criticized that, in practice, rewards are difficult to define precisely. To address this, proponents may argue that, using the free-energy principle, we can define intelligence as maximizing internal reward, with the reward function defined as the minimization of surprise. While this addresses a portion of the proposed problems, internal rewards lack testability and provide only explanatory power for observations rather than enabling direct measurement.

To avoid compromising the fruitfulness of this definition, we can use external rewards and assume that all test subjects, whether human or AI agents, will pursue these rewards. This way, we can fairly compare the agents' ability to optimize reward. However, this would render the definition incapable of accounting for various concepts of intelligence. For example, delayed gratification, causality from cognitive abilities, and so on. Furthermore, it would be akin to defining intelligence as solving a complex optimization problem (which is a special case of the previous definition). What distinguishes this definition is its emphasis on the importance of interaction with the environment and on receiving rewards from this interaction. This concept of interaction adds substantial value to the definition and should not be overlooked. Our next definition especially focuses on the interaction among the agent, the environment, and the goals the agent pursues.

## Section 3.4 Definition #4: Intelligence as Agent-Environment Relationship

### Section 3.4.1 Introduction

At the start of the 21st century, Hutter proposed AIXI, a rational agent framework that can predict sequences in uncertain worlds, and one of the earliest attempts to define intelligence universally, as artificial intelligence played an increasingly important role in our lives (Hutter, 2000). The AIXI framework is a definition of intelligence that is closely related to reinforcement learning and aims to maximize the expected total reward an agent receives from the environment. However, rather than optimizing reward as in the previous definition, Hutter emphasizes the agent's ability to achieve goals **across a variety of tasks and environments**. He later proposed a definition of intelligence with Legg that aligns with that idea, while collecting and examining more than 70 other definitions of intelligence (Legg & Hutter, 2007a).

The key finding from the collected definitions is that experts commonly recognize intelligence as a property that agents possess when interacting with their environment.

Intelligence is also widely believed to be related to an agent's ability to succeed or profit from some goal or objective. It depends on how the agent adapts to different goals and environments. Combining all of the above observations, Legg and Hutter define intelligence as: "**Intelligence measures an agent's ability to achieve goals in a wide range of environments.**" (Legg and Hutter, 2007, page 402)

For about 20 years, the idea that intelligence arises from an agent's interaction with the environment has been popular among researchers (Albus, 1991; Gudwin, 2002). For instance, Nakashima stated that intelligence is "the ability to process information properly in a complex environment."(Nakashima, 1998, page 57). In the work "On Defining Artificial Intelligence", Wang followed Legg and Hutter by also defining intelligence by emphasising its *adaptation* capacity, saying, "intelligence is the capacity of an information-processing system to adapt to its environment while operating with insufficient knowledge and resources"(Wang 2019, page 17). However, the work also identified several modifications, for example, extending the scope of adaptation to include not only adapting the agent to environmental constraints but also adapting the environment to the agent's desires. Additionally, the work pointed out that adaptation refers more to effort than to outcome. This means that intelligence, i.e., the ability to adapt to the environment, does not guarantee success; rather, it entails using available resources and knowledge to the fullest, given the constraints imposed by insufficient knowledge and resources.

A survey conducted by Monett and Lewis found that, among nine available definitions, Wang's was the most widely endorsed, with 56% of respondents agreeing or strongly agreeing (Monett & Lewis, 2018). The survey takers are experts, including researchers of the topic and AI engineers. This definition is among the closest to achieving consensus among researchers on how to universally define intelligence, in a way that not only evaluates human intelligence but also measures the intelligence of non-human agents.

Section 3.4.2 Environmental adaptation as a universal definition of intelligence

**Similarity to the explicandum**

Defining intelligence as environmental adaptation suffers from a problem similar to defining intelligence as IQ test scores: the scope of the definition. However, unlike IQ test scores, the problem isn't that its scope is too narrow but rather that it is too broad (Gignac & Szodorai, 2024, p. 5). Using an individual's ability to adapt as a standard for measuring intelligence extends the scope of what constitutes intelligence too far, potentially leading to the classification of simple organisms, such as bacteria and other microorganisms, as intelligent, since they also exhibit adaptive capabilities and undergo adaptive processes.

You can defend that it is not whether an organism adapts that matters, but whether it possesses the ability to "accomplish a variety of meaningful tasks in diverse environments". However, this does not help the definition from being overly broad. For example, some viruses are capable of infecting multiple hosts from different species, tardigrades can adapt to extreme environments such as freezing temperatures and high pressures, and small jumping spiders exhibit a complex prey-catching behavior that requires different movement patterns based on the type of prey they are catching (Jackson & Pollard, 1996). These are all examples of living beings that exhibit strong environmental adaptation but are usually considered to have little to no intelligence.

Beyond being overly broad in scope, defining intelligence as environmental adaptation also fails to account for the explanatory power of specific mental abilities widely regarded as belonging to intelligence. For example, the ability to plan for the future. The perspective of environmental adaptation leads us to overemphasize the subject's reaction to the present environment, since adaptation is defined relatively to an environment that is already given. Since the success criterion is evaluated by how well the agent adapts to environments, rather than by how the agent internally represents or anticipates future possibilities, the definition focuses on how the agent responds to environmental pressures at present. This significantly underestimates the importance of intelligent behavior that anticipates potential future environmental changes.

Lastly, scholars argue that this definition lacks a connection to mental attributes like cognitive abilities. For example, the migration behavior of birds is considered a successful adaptation to the environment, but it is also partly instinctual or genetically pre-programmed (Gignac and Szodorai 2024, page 6). Overall, successful adaptation is more the result of intelligence than its definition. Adaptation is something we can observe in action and infer from it the level of intelligence a subject possesses, but it lacks the explanatory power to answer questions such as: Why is the ability to imagine and reason about abstract concepts considered intelligent? Why is self-sacrificing sometimes considered intelligent? Or what is the difference between the intelligence level of AI and the intelligence level of humans if they have the same environmental adaptation abilities?

**Exactness**

We again want to draw attention to the semantic distinction between "adaptation" for humans and "adaptation" for AI agents. For living beings like humans, environmental adaptation can be understood as the pursuit of a survival advantage. If, by performing a behavior, the subject transitions to a state that is more advantageous for survival or reproduction than the previous state, we can consider this a successful adaptation and thus an intelligent behavior. This concept is also widely used in the context of animal intelligence. For example, bird migration is considered a successful adaptation that enhances survival. How can this be explained in the context of non-living, goal-oriented AI agents?

A widely used method is to observe the scope of the AI agent's learned task, using the scope's expansion as evidence of adaptation. For example, suppose a person and an AI system are simultaneously trained in Japan on a facial recognition task. If, after training, both systems correctly recognize faces in the US, the Netherlands, or Kenya, we can conclude that both systems have successfully adapted to the new environment for facial recognition. In this way, we have aligned the concepts for human and AI agents under this definition. Although constrained by the number and type of tasks tested, this method of evaluating adaptation can yield high exactness.

**Fruitfulness**

Proponents of defining intelligence as environmental adaptation generally hold that a subject's level of intelligence varies with the environment in which it is situated. For a subject, different environments and tasks will yield different levels of aptitude; therefore, to assess a subject's performance across all environments, it is necessary to test it in each environment individually. Some scholars refer to these environments as "umwelts" and claim that intelligence possesses umwelt-dependent properties. We can intuitively observe this property from school teachers. A history teacher, while possessing high intelligence in discussing historical contexts, may not demonstrate the same ability when answering questions about physics, and vice versa.

While this is an empirical and logical observation, when comparing levels of intelligence, we typically seek not only umwelt-dependent analyses but also an overall measure of intelligence, or a subject's generalization ability. The solution proposed by the environmental adaptation proponents is to sum up all umwelt-related test results: "To give a scalar value for intelligence, results over all umwelts must be summed together." (Gamez, 2025). However, both environments and tasks are not exhaustive. This means that it is impossible to list all combinations of environments and tasks exhaustively, and we cannot measure abilities in unseen environments or tasks. If we cannot calculate and sum the adaptation scores for every possible umwelt, there will inevitably be bias between the test environments and the real environment, which will hinder the testability of the definition.

**Simplicity**

We noted that this definition is overly broad, encompassing many living beings and behaviors that we typically do not consider intelligent. This definition is also very good in its simplicity, so I believe it is worth examining whether sacrificing a little of simplicity could refine the definition to a more precise scope. In other words, we can add premises and conditionals that may help with the definition's other aspects. For example, only successful adaptations based on an individual's beliefs and desires can be considered successful adaptations. This way, we not only require the subjects to have beliefs and desires to be intelligent, but we also provide connections to mental attributes, making the explicatum much more similar to the explicandum. However, by doing so, we also reduced the exactness and fruitfulness with respect to testability, making this attempt to extend the definition unsatisfactory.

Section 3.4.3 Conclusion

The definition of intelligence as the ability of an agent to adapt to the environment is a widely researched perspective and currently among the most commonly agreed-upon definitions between experts. Defining intelligence in terms of adaptation challenges the formal, anthropocentric view of intelligence, leaving room for a wide range of beings to be considered intelligent. However, it is also the overly wide scope that caused a big difference to the explicandum. Other differences to the explicandum, such as the inability to explain how making decisions about future events can be considered intelligent, also lowered the overall satisfaction.

Regarding the relationship between other definitions of intelligence, we are able to argue that defining intelligence as reward optimization can be inclusive of this definition. This can be done by claiming that optimizing rewards from the environment can be treated as a type of environmental adaptation. Other definitions are difficult to reconcile with this definition, as most focus on mental attributes, such as learning efficiency, for which this definition lacks explanatory power.

## Section 3.5 Definition #5: Intelligence as Learning Efficiency

### Section 3.5.1 Introduction

Some researchers believe that to obtain a fair comparison between agents, we must first ensure that all agents have the same amount of prior knowledge and that there are no hardware differences, including differences in computing power. For AI agents, this means that the only difference lies in the algorithmic structure, which can be compared by evaluating a task's learning efficiency under the assumption of unlimited data and computational resources (Chollet, 2019). It is widely recognized that the volume of human-generated data is increasing rapidly and that, in the near future, virtually unlimited data will be accessible to anyone via the internet. As the internet lowers the barrier for accessing information, the same could eventually apply to human intelligence as well. Research has shown that the way the internet provides information has already changed how the human brain works in terms of memory processes and social cognition, so one can argue that, in the future, having unlimited prior knowledge is not impossible with internet access (Firth et al., 2019).

Under the assumption of unlimited data, computing resources, and time, researchers define intelligence as the efficiency of skill acquisition, specifically the efficiency with which a task is learned. In simpler terms, it means that a subject's intelligence level is its learning efficiency. This concept of intelligence as the efficiency of learning is also evident in educational contexts, where children who are fast learners rather than walking encyclopedias are more commonly considered "gifted" (Sternberg, 1985).

In alternative terms, H. Woodrow quoted in Sternberg's "Handbook of Intelligence" that "(intelligence is) The capacity to acquire capacity"(Sternberg 2012, page 6). From an AI science perspective, this is a remarkable definition, as it shifts the focus of intelligence from outcomes (training results) to processes (training efficiency). An algorithm that converges faster is considered a better algorithm. Compared with traditional perspectives, in which results matter more but are too often controlled by the training data fed to the system, and where large corporations possess an unfair advantage, this definition ignores the unfair advantage conferred by the amount of prior data. It is finally time for AI's internal architecture, which is responsible for the training process and learning efficiency, to gain some attention.

Section 3.5.2 Learning efficiency as a universal definition of intelligence

**Similarity to the explicandum**

Unfortunately, some concept described in this definition differs fundamentally from our understanding of intelligence. The most obvious counterexample is that many studies show that children have higher learning abilities than adults (Frank et al., 2022). According to this definition, we can infer that human intelligence declines with age. This clearly contradicts the characteristics with which we are familiar regarding the intelligence explicandum. This counterexample also highlights that the premise of infinite data and computing resources is not accurate in human intelligence.

Additionally, because AI systems are currently demonstrating learning capabilities that far surpass those of humans in certain tasks, capable of learning and memorizing hundreds of thousands of data points in a short time, adopting this definition would render the level of intelligence possessed by humans far inferior to that of AI agents. This contradicts the prevailing view that AI, in general, does not yet possess intelligence that far exceeds that of humans. Therefore, I conclude that the most significant problem with this definition is its lack of similarity to the explicandum.

**Exactness**

There are also concerns regarding the exactness of this definition. Specifically, how the concept of learning differs for human and AI agents. Although the word "learning" is widely used in both contexts, several crucial differences exist between human and AI learning. For example, humans can learn from their environment through interaction, whereas most AI agents learn only during their training phase and only from processed data. Furthermore, whereas AI systems excel at supervised learning, human learning in the brain is largely unsupervised and does not rely on fixed answers. This contrast is not meant to deny the existence of unsupervised learning and reinforcement learning in AI science. Instead, even within these training methods, AI learning is typically confined to a predefined process and objective, whereas human learning remains more continuous, flexible, and environmentally interactive.

Therefore, if we aim to measure a subject's intelligence based on learning efficiency, a question arises: should we primarily test it using centralized supervised learning, or unsupervised learning—which involves interaction with the environment and lacks fixed answers—thereby leaving room for ambiguity?

Another criticism is that there are actually no fixed answers. Unlike the in-training and testing environments, many real-world problems that future AI agents might face lack correct answers. If the clarity of the issues to be addressed becomes ambiguous, will the concept of learning efficiency also lose precision? Chollet argues that the development of AI agent intelligence should not aim for a universal essence but rather for an approximation of human intelligence. This means that the behavior of human intelligence may provide the correct answer to all problems of intelligent behavior. He believes that the way to measure AI intelligence should be limited to "a specifically human-like form of intelligence, and should benchmark progress specifically against human intelligence (which is itself highly specialized)"(Chollet, 2019, p. 24). While this partially addresses the problem that the definition of learning is not sufficiently precise, given the differences between humans and AI, it also makes this viewpoint somewhat anthropocentric, sacrificing other requirements for precision.

Lastly, another issue lies in the oversimplification of the learning efficiency concept. I argue that defining intelligence as learning efficiency presupposes that subjects learn at a constant learning rate, which is a flawed assumption. We often describe the learning process as a curve, known as the learning curve. Learning curves come in varying shapes, and individuals exhibit different curves across tasks. Naturally, situations like this can arise: Student A initially learns very quickly, but their learning rate slows when they've learned 90%. In contrast, Student B learns more slowly initially, but learns faster than Student A at the 90% mark. We know that most things in the world that require learning don't have a definite endpoint, so how do we determine whether A or B has a higher level of intelligence in such cases?

**Fruitfulness**

The fruitfulness of this definition can be viewed as two-faced. On the one hand, it surely encourages AI scientists to focus on new and more powerful structures, addressing the problem of insufficient training data, which prevents them from competing with large corporations and thus hinders their proper evaluation. Research programs that focus on the architecture of AI agents rather than on actual results are more likely to produce highly efficient and high-potential AI technologies than the current results-oriented environment. Furthermore, such scientific breakthroughs will be decoupled from the amount of data available, diminishing the overwhelming advantage of large corporations.

On the other hand, the development of high-precision models, which are currently in high demand, may be slowed down. Even if a more efficient AI agent is developed, the infinite prior and infinite training data required by this definition are not readily available. Therefore, using this definition as the definition of intelligence, while potentially yielding gains in AI science, may come at the expense of societal contributions.

In contrast, both the testability and the extent to which learning efficiency in testing environments reflects real-world environments make the definition's fruitfulness satisfactory.

**Simplicity**

Finally, from a parsimony perspective, this definition relies on more complex premises than others, such as the availability of infinite resources, infinite computation time, and a constant learning rate. The question of when these premises hold is a different topic. Still, if we simply compare parsimony with the definition of environmental adaptation, it is clearly inferior, as the latter does not rely on these premises. If learning efficiency can be considered a form of environmental adaptation, then environmental adaptation without the assumption of infinite computational resources is clearly a simpler definition.

### Section 3.5.3 Conclusion

While focusing entirely on learning efficiency may appear to give a broad and fruitful definition, it ultimately suffers from the overassumption of prior knowledge and computational resources. The substantial variance in skill-gain efficiency across different stages of the learning curve or subjects also posed problems for assessing similarity to the explicandum. Finally, this definition can also be criticized for encouraging scientific improvements only in AI systems responsible for learning efficiency, rather than improving capabilities on specific, meaningful tasks.

## Section 3.6 Definition #6: Intelligence is the ability to predict

### Section 3.6.1 Introduction

Imagine you've moved to a new house. You predicted it would take you about 30 minutes to walk to school in the morning, so you left home 30 minutes before the class started. On the way, you realized you were only halfway there and 20 minutes had already passed. Knowing you would be late if you continued walking, you began running and successfully arrived on time. In recent years, scholars have paid attention to this kind of intelligent behavior. They argue that prediction and the correction of predictions in response to errors may be the nature of intelligence. This view holds that, to make accurate predictions, subjects must have internal representations of the world, and that the presence of holding such a structure internally is strongly correlated with intelligence. This view is highly appealing to AI systems that excel in predictive capabilities and have thus quickly gained significant attention across various fields.

Proponents are inspired mainly by Hohwy's proposal, which holds that the brain's most crucial function is to predict the reality around the subject, continuously verify those predictions, and minimize prediction error (Hohwy, 2013). This theory is known as predictive processing. In the previous example, there was a prediction error between the initial prediction and the actual future. Upon reaching the intermediate point, new information indicated an increased likelihood of being late, prompting you to re-predict your actions based on this new information, thus correcting the prediction error. In the theory of predictive processing, the brain uses Bayesian inference to update its probability distribution in the face of new information. New information serves as evidence that supports or modifies the predicted probability distribution, using Bayes' theorem. This is called the Bayesian update.

This behavior of minimizing prediction error is similar to the free energy principle we discussed earlier. However, compared with frameworks such as the Free Energy Principle, predictive processing employs the concepts of prediction error and Bayesian updates to more concretely explain how the brain implements a surprisal-minimizing function. Furthermore, the concept of prediction is more suitable for integration with other non-human intelligent concepts than the free-energy framework.

Since the term "prediction" is widely used in fields such as psychology, computer science, and economics, this concept has quickly attracted attention across research domains. In AI, artificial intelligence models often use inference, reasoning, or prediction to provide a judgment about the answer to a problem based on past data. Suppose the nature of intelligence follows predictive processing. It may then be possible to engineer a fully intelligent AI agent using AI prediction and the principles of predictive processing, such as Bayesian inference.

Section 3.6.2 Predictive abilities as the universal definition of intelligence

Proponents of predictive definitions observe that, while many existing theories mention elements of intelligence, such as "interacting with the environment" and "achieving goals," most fail to specify how goals are achieved or in what form interaction with the environment occurs. Taking predictive ability as the nature of intelligence effectively addresses these two issues: when interacting with the environment, the agent receives environmental information, predicts the potential consequences of its actions, chooses the optimal behavior, and influences the environment. In this process, the agent continuously updates its predictions and decisions in response to new information to minimize the discrepancy between the final prediction and the "goal". Without the prediction step, the agent will have no knowledge of what will happen if an action is taken; thus, it will effectively act randomly, contributing little to minimizing the discrepancy, and can be considered unintelligent behavior. According to Tjøstheim, this prediction error can be considered a negative reward for organisms (Tjøstheim & Stephens, 2022). Therefore, minimizing prediction error within organisms can also be understood, to some extent, as optimizing reward.

Unlike the previously mentioned reward optimization, which is based on the quality of the agent's current state, the reward here is an abstract concept constructed by the agent based on information provided by the environment and its prior knowledge. In other words, even if a subject enters a worse state after performing an action, as long as the fact of entering this state is very close to the subject's prediction, we can still say that he has achieved the optimization of his reward. For example, a comedian might hope to slip and fall at the perfect moment. Rather than having a complex or constantly changing reward function (imagine how to design such an RL robot), we would say he's trying to approximate his prediction of the "perfect moment to fall."

In non-biological-based research fields, Gamez proposed that predictive ability is equivalent to the ability to achieve goals for AI agents (Gamez, 2025, p. 5). That is, for current AI systems, we can treat their output directly as the prediction outcome. If we assume that all AI agents are designed with a single goal, the accuracy of the AI's predictions is identical to its ability to achieve that goal. However, this assumption is limited by the current "one problem, one AI" paradigm in industrial AI development. One widely agreed-upon key characteristic of intelligence, as the definition of intelligence as environmental adaptation suggests, is that intelligence must be accountable to achieve multiple goals across various environments. If we want to emphasize the goal-achieving or problem-solving ability of intelligence, in addition to the degree of achievement, we also need to consider the variety and differences of the goals. In other words, when AI agents are tasked to achieve not one, but multiple different goals, or even create new goals themselves, can we still claim that the accuracy of the AI's prediction is directly equivalent to its ability to achieve goals? This topic warrants further investigation.

## Section 3.6.3 Relationship with other definitions

Proponents of the predictive processing view of intelligence argue that prediction may be the brain's primary function; therefore, they tend to place this intellectual tradition at a lower level than other concepts used in definitions of intelligence. Because of this, predictive theory exhibits a high affinity with many definitions. It can be considered as an implementation-level definition that gives explanations for many other proposed definitions. For example, we can say that reward maximization is explanatory for intelligence because we can interpret prediction error minimization as a form of reward maximization. We can also say that the ability to optimize rewards can be explained using prediction error as the optimization target.

Furthermore, predictive theory has been shown to be highly correlated with IQ test scores (Dawson, 2025, p. 11). Gamez has claimed that, "Many intelligence tests ask the examinee to predict future items in number and shape sequences. For example, in a Raven's matrices test, two sequences of shapes are shown, and the examinee is asked to complete the third. Performance on Raven's matrices is strongly linked to general intelligence (Bartholomew 2004), so people who are better at predicting are likely to achieve higher IQ scores or g factors." (Gamez, 2025, p. 6). Giving reasons to believe that there exists some kind of relationship between prediction abilities and *g*.

Lastly, this perspective holds that intelligence involves an internal simulation (or representation) of the real world, which is utilized to enhance predictive and explanatory power. Compared to other definitions, it is also more capable of explaining how planning for the future can be considered intelligent.

However, with respect to the similarity to the explicansum, there remain important aspects of intelligence that are difficult to explain by prediction alone, making this definition controversial—for example, the importance of intelligent behaviors. A person who brings an umbrella because he predicted rain and a person who does not carry an umbrella, even if he similarly predicted the rain, will be considered having the same level of intelligence, even if their results differ tremendously. The main problem in the similarity to the explicandum and other minor issues will be discussed one by one in the following section.

**Similarity to the explicandum**

There are weaknesses that undermine the validity of this definition. To begin with, the relationship between predictive ability and intelligent behavior is weak. If intelligence requires only minimizing prediction errors, the motivation to engage in intelligent behavior will be low. In the previous example, if we treat the level of one's prediction ability as one's intelligence, it can only explain the extent of knowing that one is likely to be late, not why one should increase one's speed. In other words, predictive ability concerns only what might happen next within one's internal representational world and does not encompass spontaneous future actions. By extension, prediction systems, such as weather and earthquake forecasting, would also be considered intelligent if this definition is accepted as the universal definition of intelligence.

Additionally, this definition is criticized by scholars for failing to explain certain aspects of intelligence, such as creativity, learning abilities, and the importance of inference speed (Tjøstheim & Stephens, 2022, p. 493). More critically, although AI systems possess predictive capabilities, the Bayesian update of the predictive probability distribution emphasized by proponents of predictive processing is rarely implemented. Nevertheless, we find that the attributes of AI systems resemble those of human intelligence. We will need an explanation within the explication of why AI can appear to exhibit human intelligence without requiring Bayesian updates.

**Exactness**

There are also differences between how the term "prediction" is used in human and AI contexts that affect the exactness of this definition. Humans often use relative relationships as targets for prediction, whereas AI agents produce numerical predictions or probability distributions. For example, instead of saying that we predict the harvest to be 135% of last year, we often say things like: "the harvest will be better this year", or "it might be a better year". This is unlikely to happen for AI systems unless they are intentionally programmed to do so. An AI model is more likely to predict exact numbers than relative relationships.

Another difference is that humans make appropriate choices based on their predictions. This is because, unlike goal-oriented AI agents, humans, as living beings, combine predictions with the pursuit of their own benefits. Similarly, prediction-based decision-making is generally attributed to humans rather than to AI agents with predictive capabilities. Finally, humans can actively select the target to focus on and apply prediction, whereas current AI agents can only passively predict a predetermined objective function from the input.

**Fruitfulness**

Regarding the fruitfulness requirement, the predictive processing hypothesis yields testable predictions for human intelligence. This can be done by measuring prediction error signals and mismatch negativity in the human brain, or simply asking the subject to predict an event beforehand (Garrido et al., 2009). To interpret AI predictions, we can use explainable AI (XAI) techniques, which provide the internal representation of an AI agent's sensory inputs. Compared with metrics such as successful environmental adaptation, verifying whether a prediction is correct is relatively straightforward. This enhances the testability of the definition.

We observe that this ability to predict future environmental changes is also well-suited to comparing human and AI intelligence. This is because this ability can accurately reflect the real-world environment in the test environment. Unlike other discussed definitions that are difficult to create testing environments similar to the real world, if a prediction is generally accurate in a testing environment, it can be expected to be highly accurate in similar real-world environments as well. This is why AI systems that aim for higher prediction accuracy in training environments often achieve high accuracy in real-world environments as well.

Summarizing the two observations, we conclude that defining intelligence in terms of predictive abilities is a highly fruitful approach.

**Simplicity**

This definition also performs exceptionally well in terms of simplicity.

Conclusion

Arguing that predictive abilities can approximate the nature of intelligence provides a more fundamental definition than other definitions. Compared with other definitions, this definition explains how the brain works at a more fundamental level and argues that predictive ability, a core capability of AI, is likely the most crucial element of intelligence. This definition also explains why aspects of artificial intelligence systems may feel like aspects of living beings to some people. For example, an LLM's behavior in predicting the next word in a sentence is similar to how humans naturally predict the flow of content during conversation.

However, predictive theory is not without its flaws. Explaining the persistence of abilities such as creativity and learning remains a challenge. As discussed in the Carnap methodology section, for proponents of predictive theory, the most important consideration is which aspects to retain to yield a fruitful definition that advances scientific progress. Is it possible to extend or refine this definition to address the issues? We will attempt to achieve this in the next Chapter.

## Section 3.6.4 Further Discussion

Before moving on to our own proposed definition, I'd like to point out two common issues across definitions in the aforementioned prior research. First, I want to focus on the specific relationship between intelligence and intelligent behavior. As discussed in Section 3.6, although intelligence is often defined as an agent's ability to perform intelligent behavior, and although a third party can only perceive an agent's intelligence through the performance of intelligent behavior, the mechanisms by which intelligence enables an agent to perform intelligent behavior are rarely discussed. For example, the general factor behind IQ test scores—how does it help me survive on a deserted island? How does the ability to adapt to the environment allow me to decide how to move in the next chess move? In other words, what makes the intelligent action to be decided and taken? On the one hand, definitions such as the ability to solve complex problems and learning efficiency are satisfied by a subject merely possessing these abilities, without considering whether they can be demonstrated. On the other hand, definitions such as environmental adaptation or reward maximization appear to concern only the final outcome, not the process of acquiring and using the ability. Is there a way to explain how intelligent behavior is selected, and how this selection is fed back to the subject, preparing for the next instance of intelligent behavior?

Next, as we explored in section 3.4, is there a way to expand some flawed definitions in exchange for better performance in other requirements, sacrificing some excellent but less important simplicity? Or, could we combine two non-mutually exclusive definitions to obtain a more robust definition? I believe it is possible, provided we closely monitor for any new flaws, and I will undertake this in the next chapter.

Finally, I hope to highlight mental abilities that are not directly related to logic or the generalization of knowledge, such as creativity, social intelligence, and future planning, which are often overlooked in the definitions discussed earlier. While each of these abilities can be captured by a few distinct definitions, they are all important components of intelligence, and I aim to provide a unified definition that encompasses them.

# Chapter 4. The Extended Predictive Hypothesis

The definition I propose for a universally applicable definition of intelligence is: "**Intelligence is the ability to predict the future accurately and to benefit from that prediction.**" I call this definition the Extended Predictive Hypothesis (EPH). It is an extension of the definition, in which intelligence is understood as the ability to predict outcomes, following the predictive process framework, but also includes the capacity to benefit from predictions. In other words, an intelligent subject must accurately predict the consequences of an action and choose and perform the action that leads to a better state. The higher the level of intelligence that an agent possesses, the more accurate the prediction, and the ability to benefit from predictions will also be higher.

This definition improves on the definition of intelligence as predictive ability in two ways. First, it emphasizes that the intelligent agents' benefit must come from their predictions about the future. This can be understood as combining predictive abilities with environmental adaptation or reward optimization, while causally linking the predictive and rewarding aspects of intelligence. This means that, in this definition, the benefit does not only concern the outcome of intelligent behavior, as environmental adaptation does. Instead, the benefit is restricted to the need to be based on the subject's prediction. Therefore, intelligent subjects must possess the motivation to gain benefits in order to make predictions, and the actual benefits must follow actions that originate from the subject's predictions.

Secondly, this definition refined predictive abilities. The most significant refinement is the division of predictive abilities into two distinct types: spontaneous and reactive prediction. The difference between these two types of prediction lies primarily in two aspects: time scale and operational mechanism. Spontaneous prediction operates on a longer time scale and is a predictive ability that does not require a response to external stimuli. It conforms to the probability-distribution prediction and Bayesian update mechanisms of predictive processing theory, providing a theoretical basis for explaining abilities such as solving complex problems and learning efficiency. In human intelligence, this manifests as, for example, the ability to propose conjectures or hypotheses, plan for the future, and internalize new information.

On the other hand, reactive predictive abilities operate on a shorter timescale, providing predictive capabilities when an immediate response to external stimuli is required. Unlike spontaneous prediction, it uses (and can only use) very few inference resources and does not update the probability distribution through a Bayesian update. Therefore, it is not included in the predictive capabilities considered by predictive processing. That said, the predictive capabilities exhibited by AI are primarily reactive in nature. AI systems generate output according to their design and the input, without updating the probability distribution. In real-world scenarios, this can manifest as noticing a car approaching you and knowing you'll be hit if you don't get out of the way, or seeing a countdown and predicting the next number with ease.

Distinguishing between spontaneous and reactive prediction has another advantage: it helps narrow the scope of what qualifies as intelligent. This is because not all species possess both types of predictive abilities simultaneously. For example, fungi and plants are unlikely to possess spontaneous predictive capabilities. It is possible that current AI agents also fall into this category. We will discuss this issue further in later chapters.

I will explain why these two improvements not only significantly improve the approximation of the intelligence explicandum with respect to the definition of predictive abilities, but also enhance its exactness and fruitfulness in Section 4.5. But first, I want to start by analyzing the three main concepts of the EPH: spontaneous prediction, reactive prediction, and gainability.

## Section 4.1 Spontaneous prediction

Spontaneous prediction is the ability to simulate behaviors and their consequences in a vacuum, relying solely on the information the subject holds about the environment and themselves, without external input triggering the prediction. Some research papers on the definition of intelligence view spontaneous prediction as the only predictive ability, excluding capabilities such as object recognition or prediction that can be done by reaction, such as reading words or using simple tools (Tjøstheim and Stephens 2022, page 493; Poth et al. 2025). Analyzing this ability, we can identify several critical components of the EPH that are also observed in human prediction. First and foremost, a subject must understand the environment to make a valid prediction. This is also known as the prior knowledge requirement. If I know nothing about a city's urban structure, I certainly have no way of knowing how to get to the city museum. However, based on my prior knowledge, I can also know that asking locals or reading a map will provide sufficient information to plan a route. The EPH provides a concrete description of how knowledge contributes to intelligent behaviors and intelligence as a whole: by providing the materials needed to simulate the environment accurately.

In addition to prior knowledge, accurate spontaneous prediction also requires sound logical reasoning. It is responsible for using the "understanding of oneself and the environment" that prior knowledge provides to simulate outcomes. In plain sight, this component resembles the reasoning ability emphasized by proponents of the definition "intelligence as complex problem-solving abilities," which one might believe is an acquired ability and does not exist in beings such as human infants. However, it also has a biologically innate aspect. For example, when we see a bag with a hole and water coming out of it, we naturally know that we can help the situation by covering the hole with our hands. Another example is that infants can insert blocks into the corresponding holes of a shape sorter toy without any knowledge of the hypothesis testing process. The ability to generate heuristics about what will happen when an action is taken is fundamental.

The third component of spontaneous prediction is the ability to update predictions in response to new information from feedback. When making predictions, intelligent beings naturally include possible outcomes that are "not impossible but less likely to occur" to form a probability distribution prediction, rather than predicting a single result. For example, if you predict that a baseball team will finish the game with 5 points, even if you believe that it is improbable, the probability of them scoring 0 points or 20 points is certainly higher than zero. Sometimes, predictions suggest that a single outcome may have a sufficiently high probability that the human brain tends to ignore other possible outcomes. However, more often this probability distribution is updated, and the most likely outcome may change in response to new information. For example, if one team scored 19 points in the bottom of the eighth inning, even if it was very unlikely in the original prediction before the game, the probability of the team scoring 20 points will be significantly revised upwards. As discussed in Section 3.6, scholars suggest that intelligent agents perform Bayesian updates to update probabilities (Friston et al., 2017).

When humans experience sudden events, they may experience a brief lapse in conscious awareness while remaining fully awake. This is known as the mind-blanking effect (Fell, 2022). I believe that the mind-blanking effect provides evidence that human brains follow the probability-distribution updating function described earlier in the EPH. When a subject predicts a probability distribution, it is obvious that the distribution can include only a finite number of possible outcomes, each with a probability between 0 and 1. While most of the time the update to the prediction can be achieved by changing only the probability distribution over the same possibilities, when a sudden event that was never considered occurs, the prediction must be recalculated in light of that event. This process must take longer than regular probability updates, and during this period, the brain is unable to make other predictions, resulting in a brief lapse of consciousness.

From the aspects of intelligence they emphasize, we infer that the predictive ability we identified in our prior research is primarily limited to the outcome of combining these three components. This is why I believe that spontaneous prediction can also be understood as a narrow form of prediction, distinct from the broader sense of prediction that I propose to revise. In the broader sense of prediction, another aspect of intelligence, reactive prediction, serves as the explicatum of the accurate-prediction side of the EPH when combined with spontaneous prediction abilities.

## Section 4.2 Reactive prediction

Imagine a truck running towards you, and with a split-second decision, you duck to the side and manage to save your own life. How do you know that ducking helps? How do you know you need to get out of the way in the first place anyway?

When required to respond within a very short timeframe, humans and animals exhibit decision-making abilities that differ markedly from those described in the section on spontaneous prediction. This is another aspect of intelligence which I call reactive prediction. When events occur within a very short timeframe, such as a truck speeding towards us, or a conversation where someone is halfway through speaking, or when you see a pen roll to the edge of the table, we can, to some extent, know what is about to happen without experiencing complicated logical thinking. This ability is termed "reactive prediction" because it is an intelligent being's capacity to predict the future, but is triggered by an external stimulus. Unlike spontaneous prediction, this trigger event serves as both a mandatory starting point and a controller of the prediction horizon. This means that the reaction-prediction process terminates when the input trigger event ends. For example, when you successfully dodged a truck or caught a falling pen.

Prior knowledge remains an essential component for reactive prediction for similar reasons. The usage of prior knowledge, though, differs between the two prediction methods. Rather than logical reasoning, it is pattern recognition abilities that underpin rapid inference (Tay et al., 2016). This difference can be observed when playing reflex games, in which individuals must make rapid decisions in response to a cue. In such games, subjects often fail to switch cues after a sequence of the same correct answer. This effect, known as the "sequential effect" or "streak bias" in psychology, suggests that people expect patterns to continue when making rapid decisions (Gilovich, 1991; Livesey & Costa, 2014). In general, when required to respond quickly, intelligence relies on faster pattern recognition rather than slow logical reasoning. We observe this effect when, for example, in a fire we see the exit sign and know it indicates a path to the exit, or when we know when to jump in a volleyball game by instinct. The decisions of safety and spike positions are more based on pattern finding in what you learned or practiced, rather than logical reasoning.

Furthermore, given the short timescale of the reactive prediction process, it is unlikely that a feedback loop exists in this mechanism, since the additional time and effort required to process feedback are critical for reactive behaviors and would undermine the purpose of reactive prediction if present. Psychologists have argued that under stress or time pressure, intuitive responses may replace the entire reasoning process and become the final decision (Yu, 2016). Adapting the EPH allows us to explain this phenomenon by positing that, rather than a spontaneous prediction process, the subject engages in a reactive prediction process in situations that require short reaction times or low effort. In simpler terms, reactive prediction explains why intelligent beings tend to make single-shot, inflexible decisions under certain circumstances.

This component analysis also demonstrates the extent of bias in the intelligence of current AI agents. We observe that while current AI is highly capable of making reactive predictions by inferring from input in a very short time, it lacks capabilities in other aspects, such as spontaneous prediction. A well-trained AI model can classify objects in an image or indicate what is happening in the image. However, this lack of spontaneous predictive capabilities and the inability to continuously update possible outcomes still lead some researchers to insist that current AI agents are far from being considered intelligent (Liu et al., 2025).

From the perspective of our goal of achieving a fair evaluation and comparison of human and AI intelligence, it is clear that evaluation methods for spontaneous and reactive predictive abilities should differ, even though both are aspects of predictive capabilities. When evaluating a subject's spontaneous predictive abilities, test providers should impose a relatively loose time constraint, provide a low-stress environment, and provide the necessary prior knowledge. In contrast, when evaluating reflective prediction abilities, test providers should offer less time and only accept one-shot results.

## Section 4.3 Beneficial factor: gainability

Gainability refers to the capacity to achieve beneficial outcomes through predictive interactions with the environment. Based on our analysis at the start of Chapter 4, I argue that prediction alone is not a sufficient condition for intelligence. For example, weather forecasting systems or fortune-telling systems, while capable of predicting the future, are not considered intelligent by most people. From another perspective, as noted earlier, human perception of intelligence largely derives from observing others' behavior and consequences. Without behavior and consequences, others cannot know whether a subject possesses intelligence. Biological observations also indicate that gainability is necessary: without gaining benefits, subjects have no survival advantage. Gainability, relating to the beneficial factor, can be regarded as a part of the EPH corresponding to the "benefit from the prediction" aspect.

Some researchers, while supporting the idea that predictive abilities can serve as a good approximation of universal intelligence, deny the importance of gainability within predictive abilities (Tjøstheim & Stephens, 2022, p. 493). A large part of this exclusion is attributable to the inevitable influence of the observer's subjective opinion on the evaluation of gainability. For example, when animals lick their own wound. There are opinions on whether this act may help wounds heal faster or may instead cause infection. In this case, for an observer who believes that licking wounds promotes recovery, it is an intelligent behavior; for an observer who believes it may cause infection, it is not. I defend that the subjectivity here does not provide grounds for rejecting a definition of intelligence, since the intelligence explicandum naturally includes some subjective elements, such as individual beliefs and desires. For example, believing that the internet contains the information you need makes searching on the internet an intelligent behavior; a willingness to trade marshmallows makes the pursuit of marshmallows an intelligent behavior; and believing that money makes one happy makes the pursuit of profit an intelligent behavior.

From the perspective of biological competition, gainability is a crucial factor. Individuals with high gainability have a greater competitive advantage over others in the long run. Furthermore, we can extend the concept of gainability to account for intelligent behavior when benefits do not accrue directly to the subject. For example, if I believe that hard work will provide a better life for my family, even if I do not achieve a better state myself, this can still be considered intelligent behavior. To make this extension, we must mention two abilities that have received much attention from scholars and are deeply related to gainability: (1) proposing afforded actions (affordances), and (2) assessing and assigning value to the current state. I argue that gainability comprises these two core abilities.

I will begin by discussing the ability to propose afforded actions. An afforded action, or affordance, refers to an action from the set of possible actions an agent perceives as achievable under the limitations of its own capabilities and the environment (Gibson, 1979). Generally speaking, the more affordable actions a subject has, the more likely they are to achieve a higher benefit. In real life, this means that creative people often produce actions that unexpectedly yield positive results. This additional benefit links creativity to high intelligence. Having more afforded actions means having more possibilities. In the real world, where many problems don't have standard answers, this also means it's easier to find better actions. For example, consider two students who produce different numbers of solutions to the same mathematical problem. Obviously, the more elegant solution is likely to be found by the student who proposed ten solutions than by the student who proposed only one or two.

You might ask: Why are actions proposed as part of gainability rather than as part of predictive ability? This cognitive ability does look closely related to predictive abilities in plain sight. However, while having high predictive abilities does not guarantee a wide range of afforded actions (subjects with high predictive abilities can still come up with few afforded actions), a subject with the afforded actions advantage is sufficient to be a subject with a gainability advantage. A subject may have a strong ability to generate afforded actions even when the accuracy of its predictions is low. In such cases, we would still consider it to have high intelligence, not because its predictive abilities are superior to those of others, but because its gainability is superior.

Furthermore, when we examine the scope of active and passive predictive abilities, we find that whether our own existence is included in the prediction does not affect prediction accuracy. In other words, whether it's active or passive prediction, regardless of whether you are the main character in the prediction or not, you should be able to utilize 100% of your prediction ability. Your ability to predict the outcome of your own chess game should be completely equal to your ability to predict the outcome of a chess game on television. If you can visualize your pen rolling to the edge of the table and falling off, even when it is someone else's pen rolling to the edge, you should be able to visualize it falling as well. In the simulated world of predictive ability, your own existence has no impact on the level of predictive ability. Therefore, proposing afforded actions does not belong to predictive abilities. This also aligns with the observation that creativity is typically regarded as a relatively subjective aspect of intelligence.

However, merely identifying available actions does not satisfy the gainability criterion. Suppose a subject simulated a dozen afforded actions, and had a rough prediction of each of their outcome. Only by thoroughly examining the value that all possible actions bring to the subject can the subject make the most appropriate judgment. This leads to the second ability to discuss: judging the current state based on the environment and assigning it a value.

Those familiar with chess-playing AI or logic-based robots should be familiar with state evaluation, the process of evaluating and assigning values to the current state. But this concept is not only widely used in game agents and robotics. Definitions that emphasize the continuous interaction of intelligence with the environment almost always include this element; otherwise, the agent will interact with the environment without control over its own or its target state, thereby losing the causal basis of intelligence and behavior. For example, in the definition of environmental adaptation, the absence of an assessment of the current state's quality means the agent cannot determine whether it is better adapted to the environment than at a previous point in time or closer to achieving its goal. The absence of state self-evaluation may lead to counterintuitive conclusions, such as attributing intelligence to viruses, bacteria, and plants.

Since the state evaluation functions belong to the subject itself, we can say that if the subject values something else more than itself, then taking action based on a prediction that the something can reach a more favorable state, even if the subject itself did not, can be considered as an intelligent behavior. We often observe this phenomenon in human decision-making, sometimes resulting in self-harm for the benefit of family, religion, or the country.

As a quick summary, while I believe that predictive ability and the ability to utilize that prediction effectively can accurately approximate the nature intelligence, I am not saying that the EPH is atomic or indivisible. Instead, just as the identical quarks can form protons or neutrons, and the same protons, neutrons, and electrons can form different atoms, the EPH can be examined from three perspectives: spontaneous prediction, reactive prediction, and gainability. However, this does not mean that I believe intelligence is composed solely of logic, input, and the proposal of numerous feasible actions. Just as atoms exhibit different properties than protons and neutrons, these elements alone are insufficient to constitute intelligence or exhibit behavior that would be perceived as intelligent. Only when numerous elements work closely together, forming a subsystem, can they reveal characteristics valuable to our research interests.

Lastly, we observe that none of the above elements are restricted to biological systems. If an AI agent can recognize these individual elements and implement a system that encompasses the three aspects of EPH, I believe we have sufficient grounds to conclude that AI possesses intelligence. Furthermore, based on these three aspects, we can measure the agent's ability to perform the majority of intelligent decisions that humans can perform. Before that, we must recognize the fundamental differences between humans and AI in prediction.

## Section 4.4 Extended Predictive Hypothesis as an implementation-level definition for other definitions

In Section 3.6, we noted that predictive ability can serve as an implementation-level definition for other definitions, thereby providing a more fundamental definition that coexists with them, and characterizing intelligence as the ability to optimize rewards or adapt to the environment, grounded in the operating principle of predictive processing. As an extension of the predictive abilities definition, the EPH naturally inherits this ability. Moreover, gainability, a crucial component of the EPH, provides additional explanatory power for understanding additional aspects of intelligence. Please allow me to explain how the EPH implements other definitions one by one.

First, let us assume that the predictive process is a definition that accurately captures the nature of intelligence. Intuitively, we can offer an explanation for the prediction definition discussed in Section 3.6: it captures the part of intelligence that belongs to cognitive ability but not gainability. Therefore, it can explain some features of intelligence but cannot account for others (e.g., creativity).

Second, we can use the EPH to explain learning efficiency as follows: measuring learning efficiency is equivalent to assessing how well a subject's prediction system updates in response to information or interaction. Let's take a student preparing for an exam as an example. When they first begin working on the question set, their ability to predict correct answers may be very low. However, through feedback on gainability, for example, an improvement in the speed of answering questions, this student can judge which line of reasoning is the correct way to go and which is likely to yield incorrect answers, based on their previous answers. In essence, this revised prediction system can be understood as learned, and the efficiency of learning can be understood as the efficiency of feedback from interactions between the subject and the environment, i.e., how quickly the prediction system is updated to become accurate. Therefore, this definition yields the best prediction system only under the assumption of unlimited data and computational power, such that the final prediction system is not susceptible to bias from inaccurate data, and the subject can handle the entire process without external distraction.

From EPH's perspective, the improvement in predictive accuracy was initially limited by the rate at which the subject interacts with the world. However, under the premise of unlimited priors and unlimited computational power, the value of whether the prediction is beneficial is no longer a limitation of the system. It's like a student with an infinite set of questions, where making predictions and checking answers incurs no cost. Therefore, under this premise, for any subject capable of updating its predictions, the prediction system will eventually conform to real-world results. The difference lies only in how quickly its prediction system can update to accuracy. In other words, system-to-system comparison depends solely on learning efficiency. Furthermore, since all systems eventually become prediction-accurate through learning, the system's initial prediction accuracy is no longer critical. This is why defining intelligence in terms of learning efficiency downplays the initial prediction accuracy.

Third, we can readily explain the definition of intelligence as environmental adaptation and reward maximization through predictive processes: greater adaptability to an environment entails lower prediction error for an individual within that environment. Therefore, defining intelligence as the reduction in prediction error implies that higher intelligence is associated with greater environmental adaptability. On the other hand, for a subject, the more benefit it gains from prediction—that is, the higher its gainability—the better it is at reward maximization. We can even use gainability to explain successful adaptation. For example, someone who learns to make an igloo gains a significant survival advantage, which is both a highly gainable behavior and a successful adaptation.

As for defining intelligence as complex problem solving and IQ test scores, we can say that solving complex problems or problems in IQ tests requires the subject to predict the probability distribution of the best action to take. For paper quizzes or text questions, it means predicting the best answer. In situational problems, such as maximizing profit in a store, involve predicting the optimal course of action. Furthermore, the EPH requires that the subject benefit from the prediction to support intelligent behavior. This means that, instead of merely predicting answers, the subject must achieve a high score based on its predictions and profit from selecting the best actions while continuously updating its predictions through interaction with the environment.

The EPH can also clarify other phenomena that are difficult to explain in prior work. For example, the relationship between intelligence and intelligent behavior. The ability to perform and benefit from intelligent behavior is naturally part of intelligence, according to the EPH. However, because it largely falls within the scope of gainability, prior work that defines intelligence as cognitive abilities struggles to incorporate it into its definitions. Explaining the current development methods and landscape of AI science using the definition of intelligence is challenging, given the many definitions of intelligence. In contrast, from the perspective of the EPH, improving prediction accuracy is naturally an increase in intelligence, even if the improvement is limited to a specific domain.

Finally, I want to point out that introducing gainability as a definition of intelligence enables us to explain how learning abilities, creativity, and reaction speed relate to intelligence using prediction, which is generally difficult with previous definitions. We have already discussed that learning abilities can be viewed as successful updates to prediction accuracy that confer benefits. Creativity can actually be explained by the ability to propose affordable actions. Given equal levels of prediction accuracy and motor abilities, those who can propose more affordable actions are more likely to achieve better gainability. Therefore, if we define creativity as the degree of variance in affordable actions generated, we can infer that higher creativity yields higher intelligence through gainability. The same claim also applies to the reaction speed of a subject.

## Section 4.5 The Extended Predictive Hypothesis as the universal definition of intelligence

**Similarity to the explicandum**

Compared with the prediction-based definitions proposed, this definition not only addresses general aspects of cognitive abilities, as the predictive-process-based definitions do, but also values the **gainability** of intelligence. At a small scale, examples illustrating this definition include skilled chess players who repeatedly predict how their next move will result and then play them to gain an advantage in the game. A better player can predict the outcome of a given move more precisely, thereby increasing the likelihood of entering a winning position. I believe that this "knowing what will happen, how beneficial this outcome is to the subject itself, and receiving benefit based on that prediction" ability is a more satisfactory definition than all other definitions we discussed. Rather than being satisfied with merely obtaining an accurate prediction, it goes one step further by emphasizing the importance of the intelligent behavior underlying it.

Another good example of the extended predictive hypothesis is the fable of the crow and the pitcher. In the story, the crow threw pebbles in the water to let the water level rise, and it turned out that the rising water level did contribute to the crow successfully drinking the water. In the story, the crow is often praised for being intelligent. One way to understand this interpretation is that the crow successfully predicted the outcome of placing pebbles in the pitcher and benefited from that prediction. This indicates that intelligent predictive abilities do not merely account for intelligence observed in board games; they can also account for intelligence across broader temporal and environmental scales. However, is the EPH merely demonstrating its versatility on paper, or does it reflect the concept of intelligence we encounter in our everyday lives?

One of the most frequently used ways of describing intelligence in everyday life is through counterfactual conditionals. Counterfactual conditionals(CC) are sentences in the form of "If A had not happened, B would…", which describes the result of an assumption that an event A did not happen, given that A had already happened (or the other way around). For example, "if that lottery had won me the grand prize, I would be a billionaire by now". When describing an observation of other subjects' intelligence levels, we often use CC to emphasize the results of an action the subject took, then infer backward to the subject's intelligence level. Specifically, when we are praising someone very intelligent, we often say something like: "He is very intelligent, because had he not done A, some bad consequence would have happened." or "He is very intelligent; had he done A, some bad consequence would have happened." This tells us that when we are describing the intelligence level of other subjects through possible consequences, we are expecting that it is the intelligence of the subject that made it able to predict the outcome of its action, and the action that it took is strictly based on this prediction, bringing it to a preferable position. I will present an alternative perspective to support this point.

Another frequently used way of describing intelligence is to compare the abilities of two subjects. Consider the example "She is more intelligent than you." When comparing the intelligence level of two subjects, what is the speaker actually comparing? From the EPH's perspective, the speaker can mean two different things. One interpretation is that the speaker means that "she" knows better than "you" do about what might happen if event E occurs or action A is performed. For example, if she can predict that lighting a fire in an enclosed room filled with dust and flour in the air might result in a dust explosion, whereas you cannot, then we say that she is more intelligent than you are. This comparison is based on prediction accuracy.

There is another possibility that, instead of being better at predicting what will happen, she is better than you at benefiting from the exact prediction. For example, if both of you indicated that the olives this year would yield a good harvest, while you chose to reduce the use of olive oil until harvest season in anticipation of lower oil prices, she decided to reserve oil presses at a reasonable price. She then made a fortune by renting them when the demand is high. This indicates that she is superior in intelligence, as evidenced by her ability to benefit from the prediction.

Whether describing the intelligence of a single subject or comparing multiple subjects, the concepts of intelligence that occur in everyday life can all be explained using the EPH. The above provides evidence that the explicandum of intelligence is strongly related to the concepts of prediction and gainability in our minds, supporting this definition in terms of the similarity to the explicandum.

**Exactness**

As shown in previous sections, numerous aspects of human prediction capabilities can be captured by the EPH. Therefore, compared with prior prediction-based definitions, the EPH aligns more closely with human predictions. This means that an AI agent implementing intelligence functions based on the EPH will exhibit greater similarity in predictive performance and the benefit concept to human intelligence, thereby conferring an advantage in exactness.

**Fruitfulness**

In their definition of complex problem solving, the authors mention using simulated worlds, known as microworlds, as environments and methods for evaluating intelligence. Such an environment, capable of controlling inputs and variables, is also useful for testing intelligence defined by EPH. One reason the EPH is well-suited to testing with microworlds is that, by definition, we must verify the accuracy of the subject's predictions about the world's future and the gainability of the outcomes of the subject's actions. In the real world, such verification is time-consuming, whereas in a simulated world, we can accelerate the passage of time or even skip to the conclusion. Therefore, a well-designed testing system can perform multiple verifications in a short time, offering a high degree of testability.

For another reason, in such a testing environment, humans and AI agents can be controlled to receive the same input information and share the same prior knowledge, and it is also easy to align the test subject's target function. We can consider this a very fair testing method for both humans and AI agents. Finally, a simulated world designed based on a real-world scenario also perfectly meets the requirement that the testing environment reflects the real-world environment. Therefore, I believe this definition is also quite advantageous in terms of fruitfulness.

**Simplicity**

The only drawback is that this definition is slightly less straightforward than the definition of intelligence as predictive abilities introduced in Section 3.6. Based on our definition of simplicity, this definition is strictly weaker for simplicity than that in Sections 3.6 or 3.4. We know this because the EPH extends the definition of intelligence to include predictive abilities and gainability, whereas previous definitions considered them only partially. On the one hand, since the EPH incorporates gainability, it has lower simplicity than plain predictive abilities. On the other hand, since it emphasizes that intelligence must be grounded in predictive abilities, it also has less simplicity than plain reward optimization or environmental adaptation. However, we can also recognize the drawback of simplicity as a trade-off against greater similarity to the explicandum, exactness, and fruitfulness. Overall, I believe this tradeoff actually increases the satisfaction of this definition as a universal definition of intelligence for both humans and AI agents.

# Chapter 5. Conclusion and future directions

In this work, I argue that it is possible to establish a definition that fairly compares human and AI agent intelligence, and I set this as my research goal. I then introduce four requirements that must be closely examined to propose definitions that achieve this goal. Using these four items, we examined six definitions, drawn from different perspectives and fields of knowledge, that could serve as the definition we seek. By synthesizing the advantages of past definitions and addressing their common shortcomings, I propose a definition of intelligence based on predictive abilities and explain its advantages. This definition is called the EPH. In conclusion, I argue that the EPH is the most effective and satisfactory definition for advancing science when a universal definition of intelligence is needed. In conclusion, I hope to explore additional issues and outline directions for future research.

## Section 5.1 Discussions

How can we use this definition to evaluate the intelligence of humans and AI agents fairly?

The goal of this research is to develop a definition that enables a fair comparison of human and AI agent intelligence. So, how can the EPH help us fairly measure intelligence? And what exactly counts as a fair measurement of intelligence?

The conditions for "fairness" suggest that, firstly, all test subjects should receive the same input conditions in the same testing situation. Secondly, the evaluation benchmark should reflect the real-world for all test subjects. That is, test subjects who perform well in the testing environment should also tend to perform well in real-world environments. The use of IQ tests to assess real-world job performance has been a concern due to concerns about fairness (Richardson & Norgate, 2015).

One way to achieve this goal is to design the testing environment to closely resemble the real-world environment. Microworlds (simulated, controlled environments that function like the real world) used as a testing method for complex problem solving are a good example of testing environments that closely resemble real-world environments (Hambrick et al., 2019, p. 558). In these microworlds, we can control all inputs beyond the subject's prior knowledge, and a simulation close enough to the real world is an excellent method for evaluating performance. Furthermore, the ability to standardize motor skills across test subjects in a simulated environment is a significant advantage of using microworlds.

Now, if we accept the EPH as the most satisfactory definition of intelligence, how should we use it to compare the levels of intelligence between AI agents and humans? Using the microworlds approach, we might build environments to test AI agents in which we expect to see their usage. For example, if an AI robot that helps humans do laundry is to be evaluated, we can create an environment in which we control the amount and types of laundry, the accuracy of the weather forecast, the actual weather, and other input parameters. We can evaluate the subject's prediction accuracy by asking them when they think is a good time to hang the clothes and when is best to bring the clothes inside (they might predict based on the weather forecast or by looking at the clouds in the sky). We can also evaluate the gainability based on the clothes' wetness at the end of the simulation. This way, humans and AI agents can be tested in the same controlled environment.

What kind of scores will we see if we test current AI agents?

If the EPH is used as the definition of universal intelligence, where will the intelligence level of today's AI agents be? If we analyze it rigorously according to the EPH, we will find that we still cannot measure the intelligence of AI systems today. More specifically, AI agents that are now widely used in business cannot be considered intelligent due to their lack of spontaneous predictive ability or gainfulness across the elements of intelligence. This is true for all current goal-specific AI systems. Some AI agents employ engineering techniques to enable AI systems to make inferences without external inputs via periodic self-inferencing. However, this kind of AI system remains limited by the absence of the concept of gainability and cannot be considered truly intelligent. Perhaps this raises concern that the EPH, as a definition of intelligence, underestimates the AI agent, rendering it an unfair definition. On the contrary, people today often perceive current AI agents as lacking intelligence, merely displaying a subset of behaviors once considered achievable only by intelligent creatures.

Therefore, if the current definition is taken to mean that AI agents have sufficient intelligence to be comparable to humans, it would overestimate their intelligence. Researchers across disciplines have identified the elements they believe AI agents lack to enable fair comparisons with humans. For example, the active action-perception cycle proposed by Lake (Lake et al., 2017). Wang and Raman also claimed that self-awareness is one of the elements that distinguish human intelligence from that of AI agents (Raman et al., 2025; J. Wang, 2023).

Raman argues that self-awareness is a necessary condition for constituting an AI with a human-like mind. Wang claimed that self-awareness is an "enigmatic phenomenon unique in human intelligence" and argued that, because self-awareness is unitary and therefore uncopyable, it cannot be a non-biological property (J. Wang, 2023, p. 71).  However, is Wang's premise that "self-awareness is unitary and cannot be copied" valid? While it is true that current AI has not yet exhibited self-awareness, if the premise were true, a perfect physical copy of myself would also be unable to replicate my self-awareness. Therefore, a certain dualism must hold, and intelligence must depend on some non-physical characteristic.

I believe this claim is too radical to reject that AI agents may have the capacity to pursue intelligence.

This self is also discussed in Farina's research discussing the responsibility of artificial intelligence systems. Farina believes that a "minimal self" is required for "a sense of ownership" to exist, that is, "sense of for-me-ness"(Farina, 2022, p. 21). I argue that the sense of for-me-ness that scholars proposed as missing in AI systems is, in fact, a beneficial factor in disguise, which aligns with the concept of gainability in the EPH: "the ability to benefit from that prediction". The EPH holds that subjects must be able to clearly perceive how they interact with their environment and predict how they can improve their state through this interaction. Therefore, the claim that AI cannot be compared to human intelligence due to its lack of a sense of self can be understood as follows: scholars who argue that AI systems lack a sense of self and cannot be compared to humans may be satisfied if the AI system possesses gainability, meaning that it works in favor of a gain in its benefit. This is because an AI agent with gainability means that it can define the scope of its ownership based on the scope of benefit in its gainability function, and thus, conditionally, possess a "minimal self."

Additionally, the concept of active action-perception cycle proposed by Lake, which allows the cognitive system to "act and deliberatively explore in order to test hypotheses", is also very close to the concept of spontaneous prediction (Lake et al., 2017, p. 33). Therefore, I want to emphasize that the EPH currently does not support the claim that today's AI agents possess sufficient intelligence for a fair comparison with humans, and that this lack of elements is consistent with what scientists believe AI lacks. That is, if the future development of AI agents can address these missing elements, we will finally be able to fairly compare the intelligence levels of humans and AI agents from the perspective of universal intelligence.

## Section 5.2 Future directions

Due to the scope of this paper, there are many issues I cannot address in this work, but I believe they are highly significant for discussion.

Firstly, I argue that the EPH can fairly compare the differences in intelligence levels between humans and AI agents. As a form of intelligence, can this definition be applied to animal intelligence as well? Various indications suggest that many animals possess a certain degree of predictive ability, and their gainability-based behavior is clearly observable. If we could determine how animals predict the future and test these predictions, this definition might have substantial explanatory power regarding which kinds of animals possess intelligence and, specifically, the degree of intelligence possessed by intelligent animals.

Secondly, is this definition possible as a characterization of the nature of intelligence? This means not merely a tool for testing, measuring, and approximating the nature of intelligence, but rather a statement: "intelligence exists precisely because we internally represent the world and can use this representation to do beneficial things."

Finally, we did not explore consciousness, a concept closely related to intelligence and often discussed alongside it. According to the EPH, what is the relationship between intelligence and consciousness? If one is satisfied, will the other emerge? If one is lacking, will the other necessarily not exist? These are all questions that deserve in-depth research.

# Bibliography

Adamatzky, A. (2011). Slime mould solves maze in one pass ... assisted by gradient of chemo-attractants. In *arXiv [nlin.PS]*. arXiv. http://arxiv.org/abs/1108.4956

Albus, J. S. (1991). Outline for a theory of intelligence. *IEEE Transactions on Systems, Man, and Cybernetics*, *21*(3), 473–509.

Amodio, P., Boeckle, M., Schnell, A. K., Ostojíc, L., Fiorito, G., & Clayton, N. S. (2019). Grow smart and die young: Why did cephalopods evolve intelligence? *Trends in Ecology & Evolution*, *34*(1), 45–56.

Ang, T. L., Choolani, M., See, K. C., & Poh, K. K. (2023). The rise of artificial intelligence: addressing the impact of large language models such as ChatGPT on scientific publications. *Singapore Medical Journal*, *64*(4), 219–221.

Bal, A. P., & Or, M. (2023). Investigation of the relationship between problem-solving achievement and perceptions of students. *Pedagogical Research*, *8*(1). https://doi.org/10.29333/pr/12745

Boring, E. G. (1961). Intelligence as the tests test it. In J. J. Jenkins (Ed.), *Studies in individual differences: The search for intelligence* (Vol. 774, pp. 210–214). Appleton-Century-Crofts.

Carnap, R. (1951). Logical foundations of probability. *Pp*, *607*. https://doi.org/10.2307/2280415

Cattell, R. B. (1963). Theory of fluid and crystallized intelligence: A critical experiment. *Journal of Educational Psychology*, *54*(1), 1–22.

Chollet, F. (2019). On the measure of intelligence. In *arXiv [cs.AI]*. arXiv. http://arxiv.org/abs/1911.01547

Dawson, C. (2025). IQ, genes, and miscalibrated expectations. *Journal of Personality and Social Psychology*. https://doi.org/10.1037/pspp0000567

Deary, I. J., Strand, S., Smith, P., & Fernandes, C. (2007). Intelligence and educational achievement.

*Intelligence*, *35*(1), 13–21.

Dörner, D., & Funke, J. (2017). Complex problem solving: What it is and what it is not. *Frontiers in Psychology*, *8*, 1153.

Dreyfus, H. (1992). What computers still can't do - a critique of artificial reason.

*Https://philpapers.org › Rec › DREWCShttps://philpapers.org › Rec › DREWCS*, I – LIII,1–354.

Farina, L. (2022). Artificial intelligence systems, responsibility and agential self-awareness. In *Studies in Applied Philosophy, Epistemology and Rational Ethics* (pp. 15–25). Springer International Publishing.

Firth, J., Torous, J., Stubbs, B., Firth, J. A., Steiner, G. Z., Smith, L., Alvarez-Jimenez, M., Gleeson, J., Vancampfort, D., Armitage, C. J., & Sarris, J. (2019). The "online brain": how the Internet may be changing our cognition. *World Psychiatry: Official Journal of the World Psychiatric Association (WPA)*, *18*(2), 119–129.

Flynn, J. (1987). Massive IQ gains in 14 nations: What IQ tests really measure. *Psychological Bulletin*, *101*, 171–191.

Frank, S. M., Becker, M., Qi, A., Geiger, P., Frank, U. I., Rosedahl, L. A., Malloni, W. M., Sasaki, Y., Greenlee, M. W., & Watanabe, T. (2022). Efficient learning in children with rapid GABA boosting during and after training. *Current Biology: CB*, *32*(23), 5022–5030.e7.

Friston, K. (2010). The free-energy principle: a unified brain theory? *Nature Reviews. Neuroscience*, *11*(2), 127–138.

Friston, K., FitzGerald, T., Rigoli, F., Schwartenbeck, P., & Pezzulo, G. (2017). Active inference: A process theory. *Neural Computation*, *29*(1), 1–49.

Gamez, D. (2025). P: A universal measure of predictive intelligence. In *arXiv [cs.AI]*. arXiv. http://arxiv.org/abs/2505.24426

Ganuthula, V. R. R., & Sinha, S. (2019). The looking glass for Intelligence Quotient tests: The interplay of motivation, cognitive functioning, and affect. *Frontiers in Psychology*, *10*, 2857.

Gibson, J. (1979). The ecological approach to visual perception. *Xiv*, *332*. https://doi.org/10.2307/429816

Gignac, G. E., & Szodorai, E. T. (2024). Defining intelligence: Bridging the gap between human and artificial perspectives. *Intelligence*, *104*(101832), 101832.

Gilovich, T. (1991). How we know what isn't so: The fallibility of human reason in everyday life. *Vii*, *216*. https://psycnet.apa.org/record/1991-97937-000

Gregory, R. L., & Zangwill, O. L. (Eds.). (1987). The Oxford companion to the mind. *XVIIe Siecle*, *856*. https://psycnet.apa.org/record/1987-98496-000

Greiff, S., Holt, D. V., & Funke, J. (2013). Perspectives on problem solving in educational assessment: Analytical, interactive, and collaborative problem solving. *The Journal of Problem Solving*, *5*(2). https://doi.org/10.7771/1932-6246.1153

Griffin, A. S., & Guez, D. (2014). Innovation and problem solving: a review of common mechanisms. *Behavioural Processes*, *109 Pt B*, 121–134.

Gudwin, R. R. (2002). Evaluating intelligence: a computational semiotics perspective. *SMC 2000 Conference Proceedings. 2000 IEEE International Conference on Systems, Man and Cybernetics. "Cybernetics Evolving to Systems, Humans, Organizations, and Their Complex Interactions" (Cat. No.00CH37166)*, *3*, 2080–2085 vol.3.

Hambrick, D. Z., Burgoyne, A. P., & Altmann, E. M. (2019). Problem-Solving and Intelligence. In *The Cambridge Handbook of Intelligence* (Vol. 2, pp. 553–579). Cambridge University Press.

Hohwy, J. (2013). *The predictive mind* (Vol. 282). Oxford University Press.

Hutter, M. (2000). A theory of universal Artificial Intelligence based on algorithmic complexity. In *arXiv [cs.AI]*. arXiv. http://arxiv.org/abs/cs/0004001

Hutter, M. (2005). *Universal artificial intelligence: Sequential decisions based on algorithmic probability* [PDF]. Springer. https://doi.org/10.1007/b138233

Ivani, S. (2019). What we (should) talk about when we talk about fruitfulness. *European Journal for Philosophy of Science*, *9*(1), 4.

Jackson, R. R., & Pollard, S. D. (1996). Predatory behavior of jumping spiders. *Annual Review of Entomology*, *41*(1), 287–308.

Jaš, J. (n.d.). IQ progression of large language models. *Is.ijs.si*. https://is.ijs.si/wp-content/uploads/2025/09/AI_IN_HEALTHCARE_2025_paper_6.pdf

Kahneman, D. (2011). *Thinking, fast and slow*.

https://www.math.chalmers.se/~ulfp/Review/fastslow.pdf

Knill, D. C., & Pouget, A. (2004). The Bayesian brain: the role of uncertainty in neural coding and computation. *Trends in Neurosciences*, *27*(12), 712–719.

Lake, B. M., Ullman, T. D., Tenenbaum, J. B., & Gershman, S. J. (2017). Building machines that learn and think like people. *The Behavioral and Brain Sciences*, *40*, e253.

Lakoff, G., & Johnson, M. (1980). *Metaphors we live by*. https://www.academia.edu/download/50816110/Metaphor_we_live_by.pdf

Lea, S. E. G., Chow, P. K. Y., Leaver, L. A., & McLaren, I. P. L. (2020). Behavioral flexibility: A review, a model, and some exploratory tests. *Learning & Behavior*, *48*(1), 173–187.

Legg, S., & Hutter, M. (2007a). A Collection of Definitions of Intelligence. *Proceedings of the 2007 Conference on Advances in Artificial General Intelligence: Concepts, Architectures and Algorithms: Proceedings of the AGI Workshop 2006*, 17–24.

Legg, S., & Hutter, M. (2007b). Universal intelligence: A definition of machine intelligence. In *arXiv [cs.AI]*. arXiv. https://users.umiacs.umd.edu/~hal/courses/2016F_RL/Legg07.pdf

Lenat, D. B., & Feigenbaum, E. A. (1991). On the thresholds of knowledge. *Artificial Intelligence*, *47*(1-3), 185–250.

Liu, X. B., Fang, S., Shi, W., Wu, C.-S., Igarashi, T., & Chen, X. 'anthony'. (2025). Proactive conversational agents with inner thoughts. *Proceedings of the 2025 CHI Conference on Human Factors in Computing Systems*, 1–19.

Livesey, E. J., & Costa, D. S. J. (2014). Automaticity and conscious control in single and choice response time versions of the Perruchet effect. *Quarterly Journal of Experimental Psychology (2006)*, *67*(4), 646–664.

Logan, C. J. (2016). Behavioral flexibility and problem solving in an invasive bird. *PeerJ*, *4*(e1975), e1975.

Minsky, M. (1987). The society of mind. *The Personalist Forum*, *3*(1), 19–32.

Mischel, W. (2014). The marshmallow test: Mastering self-control. *[Produits Pharmaceutiques]*, *328*. https://psycnet.apa.org/record/2014-43233-000

Monett, D., & Lewis, C. W. P. (2018). Getting clarity by defining artificial intelligence—A survey. In *Studies in Applied Philosophy, Epistemology and Rational Ethics* (pp. 212–214). Springer International Publishing.

Nakashima, H. (1998). AI as Complex Information Processing | Minds and Machines. *Minds and Machines*, *9*(1), 57–80.

Neisser, U., Boodoo, G., Bouchard, T. J., Boykin, A. W., Brody, N., Ceci, S. J., Halpern, D. F., Loehlin, J. C., Perloff, R., Sternberg, R. J., & Urbina, S. (1996). Intelligence: Knowns and unknowns. *The American Psychologist*, *51*(2), 77–101.

Nisa, U., Shirazi, M., Saip, M. A., & Pozi, M. S. M. (2025). Agentic AI: The age of reasoning—A review. *Journal of Automation and Intelligence*. https://doi.org/10.1016/j.jai.2025.08.003

Raman, R., Kowalski, R., Achuthan, K., Iyer, A., & Nedungadi, P. (2025). Navigating artificial general intelligence development: societal, technological, ethical, and brain-inspired pathways. *Scientific Reports*, *15*(1), 8443.

Ramstead, M. J. D., Sakthivadivel, D. A. R., Heins, C., Koudahl, M., Millidge, B., Da Costa, L., Klein, B., & Friston, K. J. (2023). On Bayesian mechanics: a physics of and by beliefs. *Interface Focus*, *13*(3), 20220029.

Resnick, L. B., & Glaser, R. (1975). *Problem Solving and Intelligence*. https://eric.ed.gov/?id=ED111727

Richardson, K., & Norgate, S. H. (2015). Does IQ really predict job performance? *Applied Developmental Science*, *19*(3), 153–169.

Rosenberg, M., Zhang, T., Perona, P., & Meister, M. (2021). Mice in a labyrinth show rapid learning, sudden insight, and efficient exploration. *eLife*, *10*. https://doi.org/10.7554/eLife.66175

Roth, G., & Dicke, U. (2005). Evolution of the brain and intelligence. *Trends in Cognitive Sciences*, *9*(5), 250–257.

Schmittmann, V. D., Cramer, A. O. J., Waldorp, L. J., Epskamp, S., Kievit, R. A., & Borsboom, D. (2013). Deconstructing the construct: A network perspective on psychological phenomena. *New Ideas in Psychology*, *31*(1), 43–53.

Schnell, A. K., Amodio, P., Boeckle, M., & Clayton, N. S. (2021). How intelligent is a cephalopod? Lessons from comparative cognition: How intelligent is a cephalopod? *Biological Reviews of the*

*Cambridge Philosophical Society*, *96*(1), 162–178.

Silver, D., Singh, S., Precup, D., & Sutton, R. S. (2021). Reward is enough. *Artificial Intelligence*, *299*(103535), 103535.

Spearman, C. (1904). "general intelligence," objectively determined and measured. *The American Journal of Psychology*, *15*(2), 201.

Stadler, M., Becker, N., Gödker, M., Leutner, D., & Greiff, S. (2015). Complex problem solving and intelligence: A meta-analysis. *Intelligence*, *53*, 92–101.

Sternberg, R. J. (1985). Implicit theories of intelligence, creativity, and wisdom. *Journal of Personality and Social Psychology*, *49*(3), 607–627.

Sternberg, R. J. (Ed.). (2012). *Handbook of Intelligence* (Vol. 677). Cambridge University Press. https://doi.org/10.1017/cbo9780511807947

Sternberg, R. J. (2019). A theory of adaptive intelligence and its relation to general intelligence. *Journal of Intelligence*, *7*(4), 23.

Sternberg, R. J., & Grigorenko, E. L. (2003). Teaching for successful intelligence: Principles, procedures, and practices. *Journal for the Education of the Gifted*, *27*(2-3), 207–228.

Stevens, S. S. (1935). The operational basis of psychology. *The American Journal of Psychology*, *47*(2), 323.

Tay, S. W., Ryan, P., & Ryan, C. A. (2016). Systems 1 and 2 thinking processes and cognitive reflection testing in medical students. *Canadian Medical Education Journal*, *7*(2), e97–e103.

Tjøstheim, T. A., & Stephens, A. (2022). Intelligence is as accurate prediction. *Review of Philosophy and Psychology*, *13*(2), 475–499.

Vamplew, P., Smith, B. J., Kallstrom, J., Ramos, G., Radulescu, R., Roijers, D. M., Hayes, C. F., Heintz, F., Mannion, P., Libin, P. J. K., Dazeley, R., & Foale, C. (2021). Scalar reward is not enough: A response to Silver, Singh, Precup and Sutton (2021). In *arXiv [cs.AI]*. arXiv. http://arxiv.org/abs/2112.15422

Van der Maas, H., Kan, K.-J., & Borsboom, D. (2014). Intelligence Is What the Intelligence Test Measures. Seriously. *Journal of Intelligence*, *2*(1), 12–15.

Wang, J. (2023). Self-awareness, a singularity of AI. *Philosophy Study*, *13*(2), 68–77.

Wang, P. (2019). On defining artificial intelligence. *Journal of Artificial General Intelligence*, *10*(2), 1–37.

Wenke, D., Frensch, P. A., & Funke, J. (2004). Complex problem solving and intelligence: Empirical relation and causal direction. In R. J. Sternberg & J. E. Pretz (Eds.), *Cognition and Intelligence* (pp. 160–187). Cambridge University Press.

Wiese, W. (2024). Artificial consciousness: a perspective from the free energy principle. *Philosophical Studies*, *181*(8), 1947–1970.

Williams, S., & Huckle, J. (2024). Easy problems that LLMs get wrong. In *arXiv [cs.AI]*. arXiv. http://arxiv.org/abs/2405.19616

Yu, R. (2016). Stress potentiates decision biases: A stress induced deliberation-to-intuition (SIDI) model. *Neurobiology of Stress*, *3*, 83–95.